\documentclass{article}

\usepackage{arxiv}
\usepackage[T1]{fontenc}    
\usepackage{amsmath, amssymb, amsfonts}
\usepackage{newtxtext,newtxmath} 
\usepackage{hyperref}       
\usepackage{url}            
\usepackage{booktabs}       
\usepackage{nicefrac}       
\usepackage{microtype}      
\usepackage{graphicx}
\usepackage{doi}
\usepackage{bm}
\usepackage{multirow}
\usepackage{todonotes}
\usepackage{subcaption}
\usepackage[backend=biber,style=apa]{biblatex}
\title{OOD Detection with \textbf{immature} Models}
\author{
  Behrooz Montazeran \\
  Computer Vision and Learning Lab \\
  University of Heidelberg \\
  \texttt{behrooz.montazeran@stud.uni-heidelberg.de} \\
  \And
  Ullrich Köthe \\
  Interdisciplinary Center for Scientific Computing \\
  University of Heidelberg \\
  \texttt{ullrich.koethe@iwr.uni-heidelberg.de} \\
}

\date{} 

\begin{document}
\maketitle

\begin{abstract}
    Likelihood-based deep generative models (DGMs) have gained significant attention for their ability to approximate the distributions of high-dimensional data. However, these models lack a performance guarantee in assigning higher likelihood values to in-distribution (ID) inputs—data the models are trained on—compared to out-of-distribution (OOD) inputs. This counter-intuitive behaviour is particularly pronounced when ID inputs are more complex than OOD data points. One potential approach to address this challenge involves leveraging the gradient of a data point with respect to the parameters of the DGMs. A recent OOD detection framework proposed estimating the joint density of layer-wise gradient norms for a given data point as a model-agnostic method, demonstrating superior performance compared to the Typicality Test across likelihood-based DGMs and image dataset pairs. In particular, most existing methods presuppose access to fully converged models, the training of which is both time-intensive and computationally demanding. In this work, we demonstrate that using immature models—stopped at early stages of training—can mostly achieve equivalent or even superior results on this downstream task compared to mature models capable of generating high-quality samples that closely resemble ID data. This novel finding enhances our understanding of how DGMs learn the distribution of ID data and highlights the potential of leveraging partially trained models for downstream tasks. Furthermore, we offer a possible explanation for this unexpected behaviour through the concept of support overlap. The source code for the implementation is available in GitHub \emph{repository}\footnote{\url{https://github.com/BehroozMontazeran/ood_detection}}.
\end{abstract}

\section{Introduction}
\label{sec:Introduction}
\begin{figure}[ht!]
    \centering
    \begin{subfigure}[t]{0.49\textwidth}
        \centering
        \includegraphics[width=\textwidth]{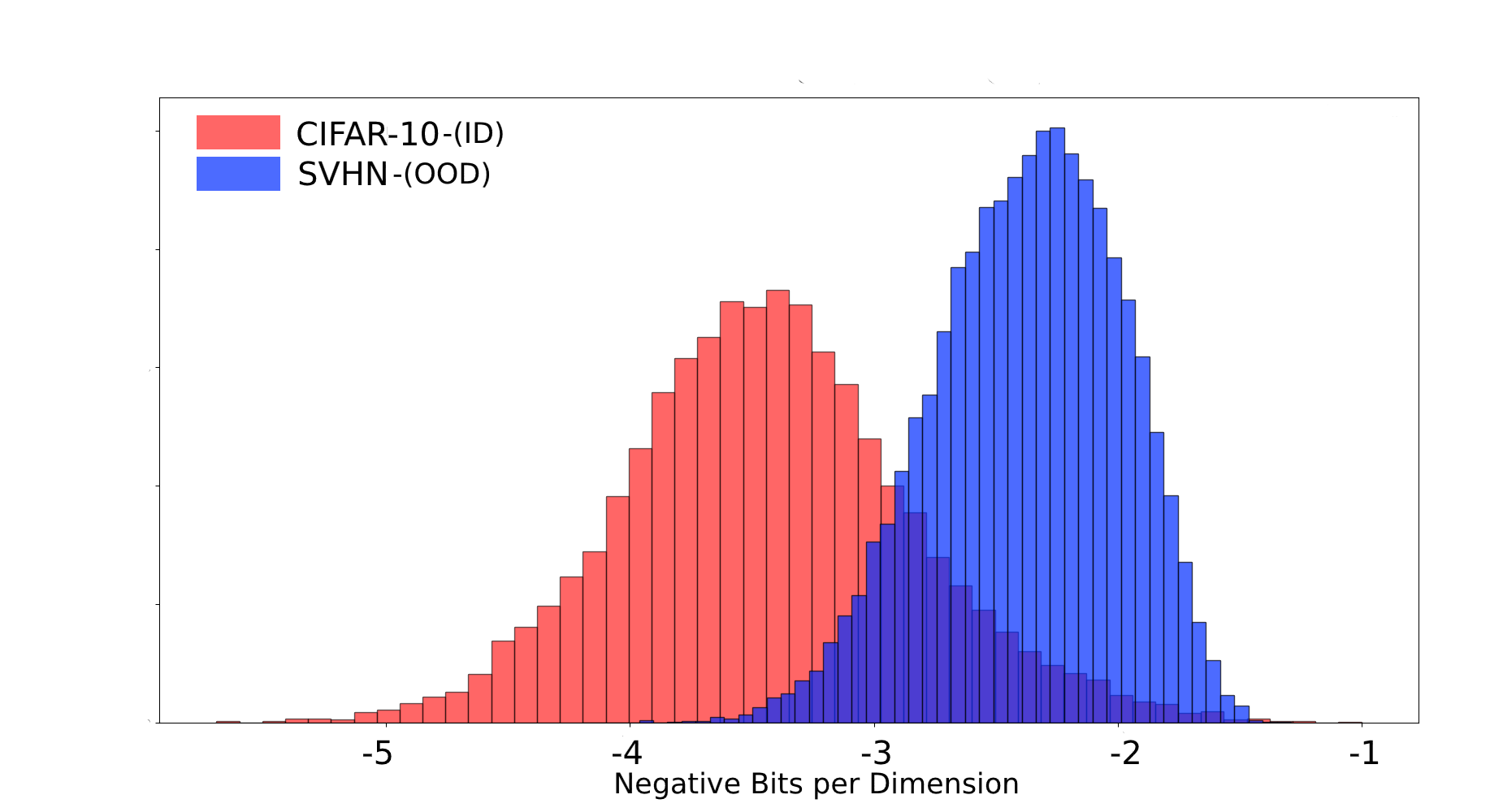}
    \end{subfigure}
    \hfill
    \begin{subfigure}[t]{0.49\textwidth}
        \centering
        \includegraphics[width=\textwidth]{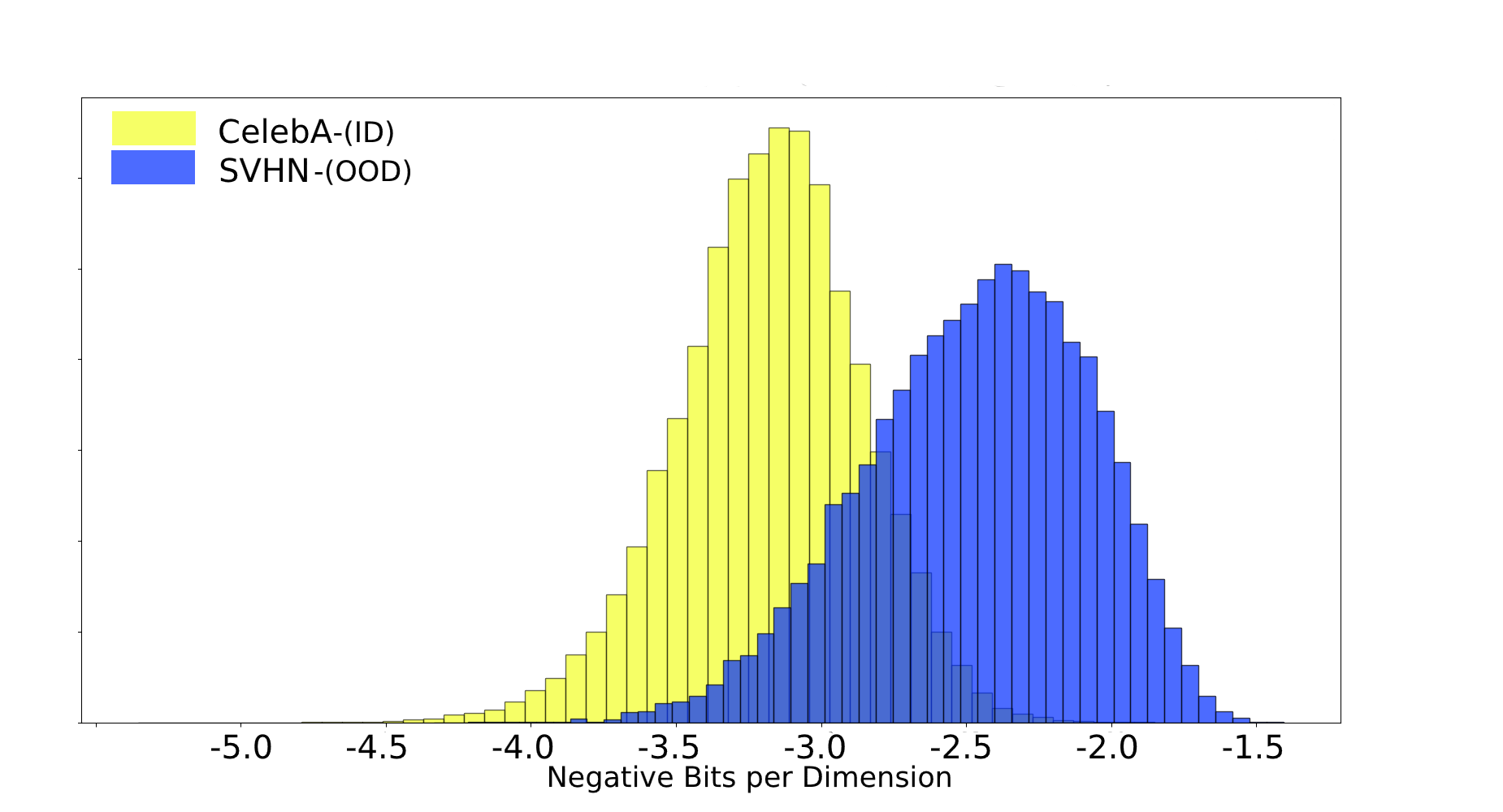}
    \end{subfigure}
    \caption{\textbf{\emph{Visualization of the anomalous behaviour in density-based generative models (GLOW)}}\\[0.5em]Despite training the model on (left) CIFAR-10 and (right) CelebA as in-distribution (ID) datasets, the model assigns higher likelihoods (higher negative bits per dimension values) to OOD samples from SVHN. This surprising observation is especially pronounced when the complexity of the ID dataset is higher than that of the OOD dataset, highlighting a key limitation of likelihood-based OOD detection in deep generative models. Further comparisons of this phenomenon are illustrated in \S~\ref{sec:NLL_as_OOD}}
    \label{fig:nnl}
\end{figure}

\begin{figure}[ht!]
    \centering
    \begin{subfigure}[t]{0.49\textwidth}
        \centering
        \includegraphics[width=\textwidth]{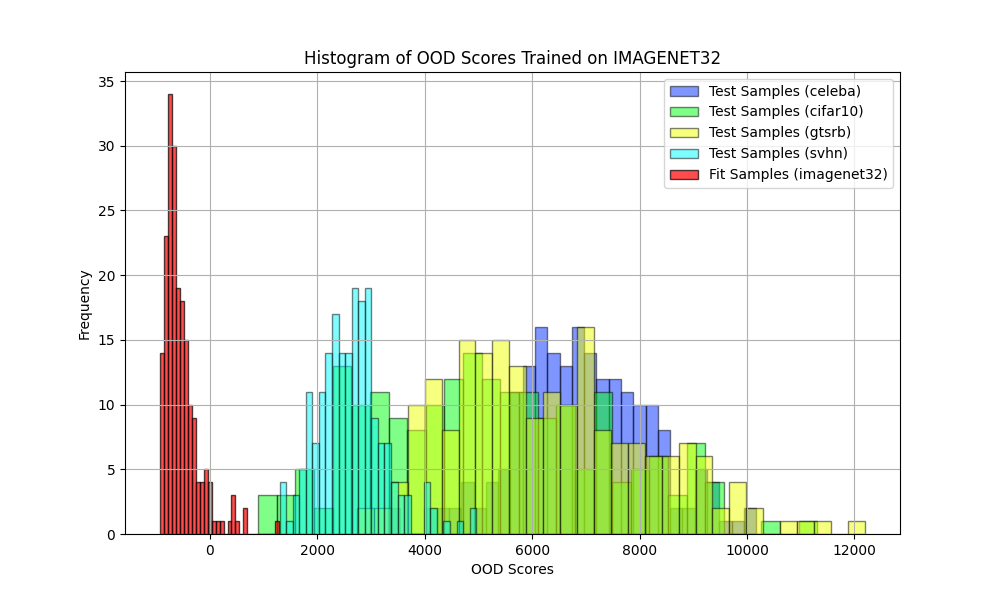}
    \end{subfigure}
    \hfill
    \begin{subfigure}[t]{0.49\textwidth}
        \centering
        \includegraphics[width=\textwidth]{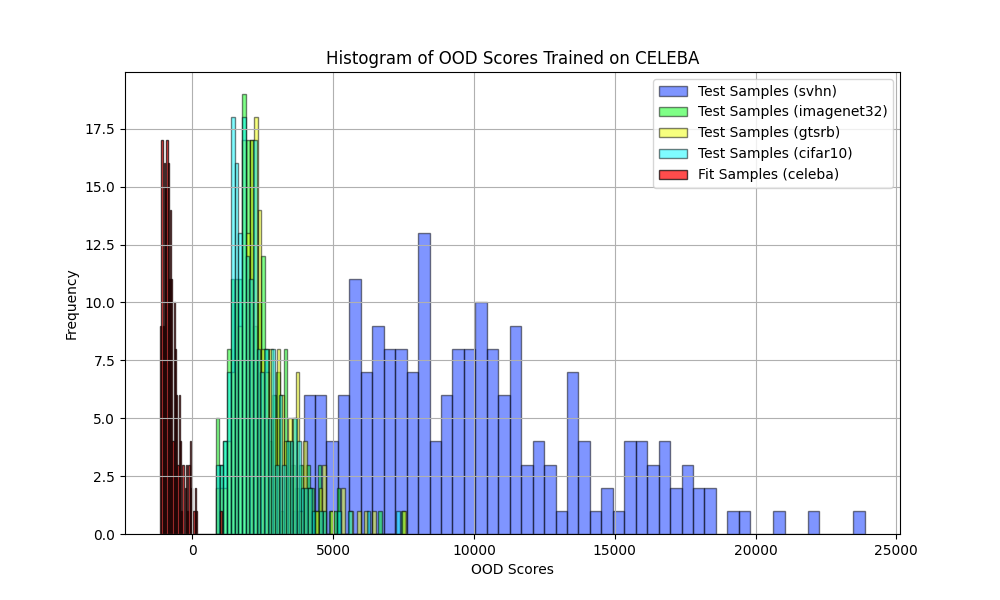}
    \end{subfigure}
    \caption{\textbf{\emph{Layer-wise gradient-based OOD scoring effectively separates ID and OOD samples.}}\\[0.5em]The GLOW model was trained on two ID datasets: (left) ImageNet32 and (right) CelebA, while tested against four OOD datasets from : SVHN, GTSRB, CIFAR-10, CelebA and ImageNet32. The gradient values are demonstrating variability across layers and the distinct separation between ID and OOD data distributions. The scoring function 
    $S_{\bm{\theta}^{(l)}}(\bm{x}_{b=\{1,5\}}) = \log \big\{ \big\| \nabla_{\bm{\theta}^{(l)}} \big( \sum_b \ell(\bm{x}_b) \big) \big\|_2^2 \big\}$
    is computed using $b=5$, indicating that each score is calculated using a batch of five random samples. The near-perfect separation observed between ID and OOD samples highlights the effectiveness of this method. Additional results with varying batch sizes (e.g., $b=1$ and $b=5$) and other IDs are detailed in \S~\ref{sec:Additional_Results}.}
    \label{fig:layerwise_scores}
\end{figure}

The application of machine learning models in high-stakes scenarios, where safety and reliability are paramount, has sparked significant debate (\cite{goodman2017european,cluzeau2020concepts,zhang2021understanding}), particularly in domains such as finance (\cite{chen2024generalized}), autonomous driving (\cite{voronin2024enhancing}), and medical diagnostics (\cite{varoquaux2022machine,su2024machine}). These models, despite their strong performance on data similar to their training distribution, often exhibit overly confident or erroneous behaviour when encountering out-of-distribution (OOD) inputs (\cite{wei2022mitigating, tony2012isolation}).
In this work, we explore OOD detection using likelihood-based deep generative models (DGMs) in an unsupervised method, especially flow-based models (\cite{kingma2018glow}), which are designed to estimate the underlying probability density of the observed data. Density estimation techniques do not assume the existence of an anomaly distribution at training time. These models, trained with objectives such as maximum likelihood, aim to maximize the likelihood of the training data, inherently normalizing the probability densities across the data distribution. This normalization suggests that OOD samples should theoretically yield lower likelihoods compared to in-distribution data (\cite{bishop1994novelty}). Notable likelihood-based DGMs, including normalizing flows (NFs) (\cite{papamakarios2021normalizing}), diffusion models (DMs) (\cite{sohl-dickstein2015deep, ho2020denoising}), variational autoencoders (VAEs) (\cite{kingma2014autoencoding,rezende2014stochastic}), autoregressive models (ARMs) (\cite{oord2016wavenet,oord2016pixel}), and energy-based models (EBMs) (\cite{lecun2006tutorial,xie2016theory}), have demonstrated remarkable capabilities, particularly in generating high-quality and highly realistic images. Building on this success, a natural approach for OOD detection is to assess the likelihood of a given sample under the trained model and classify it based on a threshold (\cite{ren2019likelihood, nalisnick2019detecting, kamkari2024geometric, xiao2020likelihood}).
However, despite their theoretical appeal, recent studies reveal a surprising inconsistency in this approach (see Fig.\ref{fig:nnl}, showing performance inconsistencies). These models, which approximate data log-likelihoods or their surrogates, often assign higher likelihoods to OOD samples than to those from the training distribution (\cite{choi2019waic,nalisnick2019deep}), especially when ID inputs are more complex than the OOD sample points. This paradoxical phenomenon, observed across various probabilistic DGMs, highlights a fundamental challenge in using likelihood as a standalone metric for OOD detection. Recent studies (\cite{choi2021robust, dauncey2023on, dauncey2024approximations}) proposed using layer-wise gradient norms as a more effective indicator for OOD detection. This approach is based on the intuitive observation that OOD data typically produce larger gradient norms compared to ID data when the model undergoes a single-step back-propagation, starting with parameters optimized on the ID dataset, as illustrated in (Fig.\ref{fig:layerwise_scores}). In essence, this method evaluates how much meaningful signal can be derived solely from the gradients, offering a fresh perspective on OOD detection.

Authors (\cite{choi2021robust,dauncey2024approximations}), along with studies such as (\cite{hendrycks2018deep, nalisnick2019detecting, xiao2020likelihood,havtorn2021hierarchical, bergamin2022model, Haroush2021statistical}) and more operate under the assumption that the parameters of a trained model are optimal, implying that fully trained models are inherently superior for downstream tasks. However, \textbf{in this contribution} we challenge this assumption by demonstrating that a layer-wise score-based OOD detection method achieves superior results when applied to partially trained models. Specifically, our findings suggest that fully optimized models may not always be the best choice for tasks like OOD detection, especially when the ID data points are less complex than the OODs samples.

To substantiate our claim, we utilized the GLOW model \cite{kingma2018glow}, evaluating its performance at different stages of training. The results, presented in Tables \ref{tab:auc_results_three_channel} and \ref{tab:auc_results_one_channel}, reveal that full convergence is unnecessary when the objective is limited to OOD detection. In fact, partially trained models not only match but mostly exceed the performance of fully trained models, especially when the OOD samples are semantically dissimilar to the in-distribution data.

We attribute this phenomenon to the evolving gap (or overlap) between the support of ID and OOD data as training progresses. Specifically, when a score-based method is used to detect OOD samples, the gap between the histograms of layer-wise gradient norms of samples under the parameters of DGMs widens progressively during training, which can impose an additional computational burden on such downstream tasks. Conversely, in some cases, this gap transitions to an overlap, or an existing overlap becomes more pronounced at the later stages of convergence. When likelihood alone, or in combination with other measures, is used as an OOD score, partial training reduces the extent of this overlap, thereby enhancing the model's ability to distinguish OOD samples. These phenomena, illustrated in \S~\ref{sec:partial_vs_full}, highlight the potential benefits of partial training. This insight offers a fresh perspective on OOD detection, emphasizing the value of leveraging partially trained models for efficient and effective performance in specialized tasks.

\section{Methods to detect OOD data (Related work)}
\subsection{Density Scoring Methods}
\label{sec:density_scoring_methods}
This section provides an in-depth examination of recent advancements in OOD detection methodologies, focusing on their underlying principles, strengths, and limitations. First, we explore likelihood-based scoring methods, tracing their evolution, and analysing their primary bottlenecks, including their dependence on accurate data representation and susceptibility to misleading likelihood estimates for OOD samples. Following this, we delve into gradient-based methods, presenting them as a promising alternative. These methods leverage layer-wise gradient information to address the challenges inherent in likelihood-based approaches, offering improved robustness and interpretability in OOD detection.

A foundational strategy for OOD detection, applicable in both labelled and unlabelled settings, involves learning a density model \(\mathcal{M}\) that approximates the true distribution \(p^* (\mathcal{X}_{\text{ID}})\) of training inputs \(x_{\text{ID}} \in \mathcal{X}_{\text{ID}}\) (\cite{bishop1994novelty}). The underlying assumption is that if the approximation is sufficiently accurate, such that \(p(x_{\text{ID}} | \mathcal{M}) \approx p^* (x_{\text{ID}})\), then OOD inputs should yield low likelihood scores under \(\mathcal{M}\). While this approach was historically considered infeasible for high-dimensional data like images or audio due to the challenge of constructing high-quality density models, recent advancements in generative modelling have made significant progress in this area.

Autoregressive and invertible models, such as PixelCNN++ (\cite{salimans2017pixelcnn}) and Glow (\cite{kingma2018glow}), have shown the capability to approximate complex data distributions with remarkable accuracy. These models fit a generative distribution \(p_\theta (x_{\text{ID}})\) to input data and evaluate the likelihood of new inputs to differentiate ID and OOD samples. However, studies by (\cite{choi2019waic}) and (\cite{nalisnick2019deep}) revealed that using model likelihood as an OOD score often results in counter-intuitive behaviour when datasets of varying complexity are paired, undermining its reliability (see our replication of their results in Fig. \ref{fig:nnl} and \S~\ref{sec:NLL_as_OOD}.

To address these limitations, various techniques have been proposed. The \textbf{Typicality Test}, introduced by (\cite{nalisnick2019deep}), uses the concept of a "typical set" from information theory (\cite{shannon1948mathematical}). It evaluates whether test data likely belongs to the generative model’s distribution by computing a test statistic \(\hat{\epsilon}\), derived from negative log-likelihoods of samples. This test operates as an omnibus goodness-of-fit (GoF) measure (\cite{eubank1992asymptotic}), ensuring that ID data falls within a high-probability region.

\textbf{Likelihood Ratio} Methods proposed by (\cite{ren2019likelihood}) isolate semantic information from background statistics in data. This approach trains a background model by adding perturbations to corrupt semantic structure, ensuring it captures only population-level background characteristics. Using the ratio of likelihoods from the original model and the background model, this method distinguishes ID data from OOD data, providing a background contrastive score for detection.

\textbf{Generative Ensembles}, suggested by (\cite{choi2019waic}), combine likelihood-based models with predictive uncertainty estimation. The method uses ensemble variance and the Watanabe-Akaike Information Criterion (WAIC) to adjust likelihood estimates, penalizing high variance across independently trained models. This approach distinguishes anomalies by incorporating Bayesian posterior approximations and leveraging variance-based metrics to highlight distributional differences.

\textbf{Hierarchical Log-Likelihood Ratios}, explored by (\cite{schirrmeister2020advances, havtorn2021hierarchical}), address the dominance of low-level features in likelihood-based methods. By training generative models on both general and specific ID datasets, the ratio of their likelihoods isolates semantic information, mitigating biases from low-level features. Additionally, an outlier loss function, incorporating temperature-scaled log-likelihood ratios, enhances the robustness of OOD detection. However, finding a comprehensive general dataset in practice is not feasible.

\textbf{Density of States Estimation (DoSE)}, developed by (\cite{morningstar2021density}), introduces a statistical mechanics-inspired technique. It estimates the density of states from summary statistics of pre-trained generative models. By fitting distributions to these statistics on ID data, DoSE identifies OOD samples as those with low density support, providing a novel perspective on OOD detection.

(\cite{kamkari2024geometric}) introduced a technique leveraging \textbf{Local Intrinsic Dimension (LID)} (\cite{bengio2012unsupervised, pope2021intrinsic}) to address the paradoxical behaviour of generative models assigning high density but low probability mass to OOD regions. Using LID estimates alongside the model likelihood, the method classifies data points into three cases: (1) low likelihood, (2) high likelihood with low LID, and (3) high likelihood and high LID. In the first two cases, the assigned probability mass is negligible, indicating OOD classification, while the third case suggests the data point is ID. This approach demonstrates the significance of considering both density and LID for reliable OOD detection.

A critical challenge in all density-based methods is ensuring \emph{representation invariance} (\cite{lan2021perfect, serran2019input}), which is often not guaranteed except in methods involving density ratios. However, even ratio-based approaches face practical hurdles, such as identifying an appropriate global dataset to serve as the denominator—a task that is often unattainable. This lack of invariance highlights the susceptibility of density-based methods to arbitrary manipulations, such as using different data representations (e.g., RGB or HSV colour spaces), which can introduce subjectivity and compromise reliability.

Our method distinguishes itself from existing likelihood-based OOD score methods by utilizing the layer-wise gradient of parameters, a feature absent in traditional approaches. Unlike \cite{dauncey2024approximations} and \cite{choi2021robust}, which rely on statistics derived from fully trained models, our approach leverages partially trained models, demonstrating that these can perform better for OOD detection. Additionally, while \cite{zhang2021understanding} focuses on explaining the performance of immature models with histogram of support overlap, our method provides a novel perspective by analysing the gap between the OOD scores logarithm gradient $L^2$-norms $S_{\bm{\theta}^{(l)}}(\bm{x})$, offering a new explanation for why partially trained models are more effective.

\subsection{Gradient Scoring (Score-based) Method}
\label{sec:Gradient_scoring}

Gradient scoring methods are renowned for representation-invariant techniques such that the choice of representation does not dramatically influence the output of the OOD detection. At a conceptual level, new sample data \( \bm{x} \) at test time can be categorized based on the variations observed in the parameters of a corresponding trained model. Minimal alterations in optimal trained parameters \( \hat{\bm{\theta}} \) typically indicate that the input data originates from the ID (\( \bm{x} \sim p^* \)). Conversely, substantial changes in the parameters are indicative of input samples belonging to OOD (\( \bm{x} \not\sim p^* \)). 

To model this formally, we represent the layer-wise parameters as \( \hat{\bm{\theta}}^{(l)} \), where \( l \) denotes the layer index. For a new input sample \( \bm{x} \), the change in parameters after incorporating \( \bm{x} \), at layer \( l \), can be expressed in terms of the likelihood \( p(\hat{\bm{\theta}}_{\bm{x}}^{(l)} \mid \cdot) \). Define the change in parameters for layer \( l \) as:

\begin{align}
\Delta_{\bm{\theta}}^{(l)}(\bm{x}) &= \log p(\hat{\bm{\theta}}_{\bm{x}}^{(l)} \mid \bm{x}) - \log p(\hat{\bm{\theta}}^{(l)} \mid \bm{x}). 
\label{eq:delta}
\end{align}

Based on the magnitude of \( \Delta^{(l)} \), \( \bm{x} \) is classified  as:

\begin{align}
\begin{cases} 
\text{ID}, & \Delta \leq \varepsilon, \\
\text{OOD}, & \Delta > \varepsilon. 
\end{cases} \label{eq:classification}
\end{align}

While this approach is theoretically sound, retraining the model with new data points to measure parameter changes directly is computationally prohibitive. To address this inefficiency, (\cite{choi2021robust}) propose using a second-order analysis to approximate \( \Delta^{(l)} \):

\begin{align}
\Delta_{\bm{\theta}}^{(l)}(\bm{x}) &\approx 
\big[ \nabla_{\bm{\theta}^{(l)}} \log p(\bm{x} \mid \bm{\theta}^{(l)}) \big]^\top F(\bm{\theta})^{-1} \nabla_{\bm{\theta}^{(l)}} \log p(\bm{x} \mid \bm{\theta}^{(l)}), \label{eq:approx_delta}
\end{align}

where \( F(\bm{\theta}) \) is the Fisher information matrix (FIM) for \( p^*(\bm{x}_{\text{ID}}) \) evaluated at \( \bm{\theta} \). This approximation provides a layer-wise representation-invariant framework for deriving a more informative OOD score than the raw model likelihood. By leveraging the Fisher Information Metric (\cite{fisher1920mathematical}), which naturally quantifies the size of the gradient, this method achieves computational efficiency and robustly distinguishes ID from OOD samples (\cite{ly2017tutorial}).

The score function  $\big\| \nabla_{\bm{\theta}} \ell(\bm{x}) \big\|_{\text{FIM}}^2$ is given by:

\begin{align}
\nabla_{\bm{\theta}} \ell(\bm{x}) &= \nabla_{\bm{\theta}} \log p(\bm{x} \mid \bm{\theta}), \label{eq:gradient_score} \\
\big\| \nabla_{\bm{\theta}} \ell(\bm{x}) \big\|_{\text{FIM}}^2 &= 
\big[ \nabla_{\bm{\theta}} \ell(\bm{x}) \big]^\top F(\bm{\theta})^{-1} \nabla_{\bm{\theta}} \ell(\bm{x}), 
\label{eq:fim_score} \\
F(\bm{\theta}) &= \mathbb{E}_{\bm{x}_{\text{ID}} \sim p_{\bm{\theta}}} \big[ \nabla_{\bm{\theta}} \ell(\bm{x}_{\text{ID}}) \nabla_{\bm{\theta}} \ell(\bm{x}_{\text{ID}})^\top \big]. 
\label{eq:fim_matrix}
\end{align}

However, directly computing the inverse of \emph{Fisher Information Matrix} \( F(\bm{\theta}) \) is computationally impractical for large-scale DGMs due to the high dimensionality of the parameter space (\( |\bm{\theta}| \times |\bm{\theta}| \)) that exceeds millions. To address this, approximation of \( F(\bm{\theta})^{-1} \) is necessary. The following methods are commonly used for such approximation:

\begin{itemize}
    \item \textbf{Identity} (\cite{bergamin2022model}): \( F(\bm{\theta}_0) \approx \mathbb{I}, \)
    \item \textbf{Diagonal} (\cite{kirkpatrick2016overcoming}): \( F(\bm{\theta}_0) \approx \mathbb{E}_{\bm{x}_{\text{ID}} \sim p_{\bm{\theta}}} \big[ \text{diag} \big( \nabla_{\bm{\theta}} \ell(\bm{x}_{\text{ID}}) \nabla_{\bm{\theta}} \ell(\bm{x}_{\text{ID}})^\top \big) \big], \)
    \item \textbf{EKFAC} (\cite{martens2016second, george2018fast}): \( F(\bm{\theta}_0) \approx \mathbb{E}_{\bm{x}_{\text{ID}} \sim p_{\bm{\theta}}} \big[ \nabla_{\bm{\theta}} \ell(\bm{x}_{\text{ID}}) \nabla_{\bm{\theta}} \ell(\bm{x}_{\text{ID}})^\top \big] = U \Sigma U^\top, \)
    \item \textbf{EKFAC using Kronecker product} (\cite{choi2021robust}): \( \mathbb{E}_{\bm{x}_{\text{ID}} \sim p_{\bm{\theta}}} \big[ \nabla_{\bm{\theta}} \ell(\bm{x}_{\text{ID}}) \nabla_{\bm{\theta}} \ell(\bm{x}_{\text{ID}})^\top \big] \approx A \otimes B, \)
\end{itemize}

where \( U \) is an orthogonal matrix, \( \Sigma \) is a diagonal matrix, and \( A \) and \( B \) represent smaller matrices derived from (\( h \))  represents the layer's input features, and (\( \delta \)) represents the back-propagated gradients with respect to the pre-activations. 

The identity approximation simplifies the FIM-based norm \( \| \cdot \|_{\text{FIM}}^2 \) to the Euclidean norm \( \| \cdot \|_2^2 \), offering a computationally efficient alternative for evaluating gradient magnitudes. This approach is particularly useful in large-scale models, where direct computation of the FIM can be prohibitively expensive. Formally, this approximation can be expressed as:

\begin{align}
\big\| \nabla_{\bm{\theta}^{(l)}} \ell(\bm{x}) \big\|_{\text{FIM}}^2 &\approx \big\| \nabla_{\bm{\theta}^{(l)}} \ell(\bm{x}) \big\|_2^2. \label{eq:euclidean_norm}
\end{align}

(\cite{bartlett1946statistical}) demonstrated that $\big\| \nabla_{\bm{\theta}^{(l)}} \ell(\bm{x}) \big\|_{\text{FIM}}^2$ follows a chi-squared (\( \chi^2 \)) distribution with a large degrees of freedom \( |\bm{\theta}| \), where \( |\bm{\theta}| \) denotes the dimensionality of the parameter space, typically estimated via MLE. Consequently, the logarithm of this squared norm should approximately follow a normal distribution (\cite{dauncey2024approximations}). Leveraging this property, the OOD score based on the log-transformed gradient norm is defined as:

\begin{align}
S_{\bm{\theta}^{(l)}}(\bm{x}) &= \log \big\{ \| \nabla_{\bm{\theta}^{(l)}} \ell(\bm{x}) \|_2^2 \big\}. \label{eq:ood_score}
\end{align}

\section{Methodology}
\label{sec:Methodology}
In the following section, we revisit and redefine a method for OOD detection based on layer-wise gradients, as proposed by (\cite{dauncey2024approximations}), and formalized in (Equation \ref{eq:ood_score}). This approach exploits GLOW (\cite{kingma2018glow}) and demonstrates that partially trained models can deliver equivalent or superior performance compared to fully trained models in OOD detection tasks.

Our findings suggest that the superior performance of partially trained models is tied to the phenomenon of sufficient gap between ID and OOD data. As training progresses on complex datasets (e.g., ImageNet32), this gap increases but is not necessarily required for achieving optimal results. Conversely, on less complex datasets (e.g., SVHN), the gap diminishes and transitions into overlap, which reduces the effectiveness of OOD detection criteria in fully converged models. This observation is explored in greater detail in \S~\ref{sec:partial_vs_full}, where we provide a comprehensive analysis of how training dynamics influence OOD performance.

\subsection{Training the Model}
\label{sec:Training_the_model}
Unlike conventional OOD detection methods that rely on fully converged models and their associated parameter statistics as OOD metrics, we propose a novel approach applying the GLOW model. Specifically, we train the model on various three-channel colour image datasets, including SVHN, GTSRB, CIFAR-10, CelebA, and ImageNet32, as well as one-channel colour such as (MNIST, FashionMNIST, KMNIST and Omniglot), but only partially. This approach extracts parameter values from ID datasets without requiring full convergence, challenging the traditional reliance on fully optimized models (see the experimental setup in \ref{sec:Experimental_Setups}). By significantly reducing computational time and energy, this method achieves remarkable improvements in OOD detection performance, yielding an AUROC metric of about perfect across nearly all tested dataset pairs (refer to Tables \ref{tab:auc_results_three_channel} and \ref{tab:auc_results_one_channel}) and challenges the necessity of fully trained model parameters.

\subsection{Extracting OOD scores}
\label{sec:Extracting_ood_scores}

To compute the OOD scores, the method allows for two approaches: single-sample OOD detection (singleton designation) and detection based on a batch of five samples, utilizing a goodness-of-fit (GoF) test. After partially training the model, samples from the in-distribution and out-of-distribution test datasets are used for one step of back-propagation. This step is initialized with the parameters of the partially trained model and captures the gradient changes caused by the new samples. These gradient changes are evaluated layer by layer, producing layer-wise scores that reflect the divergence between ID and OOD data. The scores are then combined to compute a final OOD score, which effectively determines whether the given sample or batch of samples belongs to the ID or OOD data distribution.

Having approximated the Fisher Information Matrix (FIM) as identity, the score is expressed as:

\begin{align}
S_{\bm{\theta}^{(l)}}(\bm{x}_{b=\{1,5\}}) &= \log \big\{ \big\| \nabla_{\bm{\theta}^{(l)}} \big( \sum_b \ell(\bm{x}_b) \big) \big\|_2^2 \big\}, 
\label{eq:score_single_batch}
\end{align}

where \( \bm{x} \) is the new data sample from the test dataset, and \( b = \{1,5\} \) represents the batch size. A batch size of \( b = 1 \) corresponds to single-sample OOD detection, while \( b = 5 \) enhances performance through a goodness-of-fit test.

The computed layer-wise score \( S_{\bm{\theta}^{(l)}}(\bm{x}_{b=\{1,5\}}) \) follows an approximately normal distribution, then a natural method of combining the layer-wise \( L^2 \)-norms is by fitting normal distributions to each score independently:
\begin{align}
\mu^{(l)} &= \text{MEAN} \big(S_{\bm{\theta}^{(l)}}(\bm{x}_{b=\{1,5\}})^{(N)}\big), \label{eq:mean_layer} \\
\sigma^{(l)} &= \text{VARIANCE} \big(S_{\bm{\theta}^{(l)}}(\bm{x}_{b=\{1,5\}})^{(N)}\big), \label{eq:variance_layer}
\end{align}
where \( N \) represents the number of samples from the ID dataset used to calculate the mean and variance for each layer independently. These statistics are derived from a fit dataset that is not used during the model's training phase 

Using the layer-specific means and variances, we calculate the Gaussian negative log-likelihood as a comprehensive OOD score:

\begin{align}
S_{\bm{\theta}}(\bm{x}_{b=\{1,5\}}) &= \frac{1}{2} \sum_{l=1}^{L} \bigg( \frac{\big(S_{\bm{\theta}^{(l)}}(\bm{x}_{b=\{1,5\}}) - \mu^{(l)}\big)^2}{(\sigma^{(l)})^2 + \varepsilon} 
+ \log \big(2\pi((\sigma^{(l)})^2 + \varepsilon)\big) \bigg), \label{eq:ood_score_final}
\end{align}
where \( \mu^{(l)} \) and \( \sigma^{(l)} \) are the layer-specific mean and variance from fit dataset and $\varepsilon$ is a small constant added for numerical stability.

This formulation quantifies the deviation of log-transformed gradient magnitudes from their expected mean, normalized by variance, across all layers. By summing these normalized deviations, the final OOD score effectively integrates layer-wise gradient information to detect out-of-distribution samples. As depicted in \S~\ref{sec:Additional_Results}, this summation does not necessarily result in a normal distribution but instead produces a right-skewed distribution, which arises due to the high correlation between adjacent layers.

\section{Immature vs. Mature Model in OOD Detection}
\label{sec:partial_vs_full}

We investigate the relationship between the gap (or overlap) in histograms of OOD scores (using Equation \ref{eq:ood_score_final}) across different datasets. Then, we reference the context of support overlap as discussed in (\cite{zhang2021understanding}) to demonstrate that partial training can outperform full training in OOD detection when using log-likelihood models and their associated NLL statistics.

\begin{figure}[ht!]
    \centering
    \begin{subfigure}[t]{0.49\textwidth}
        \centering
        \includegraphics[width=\textwidth]{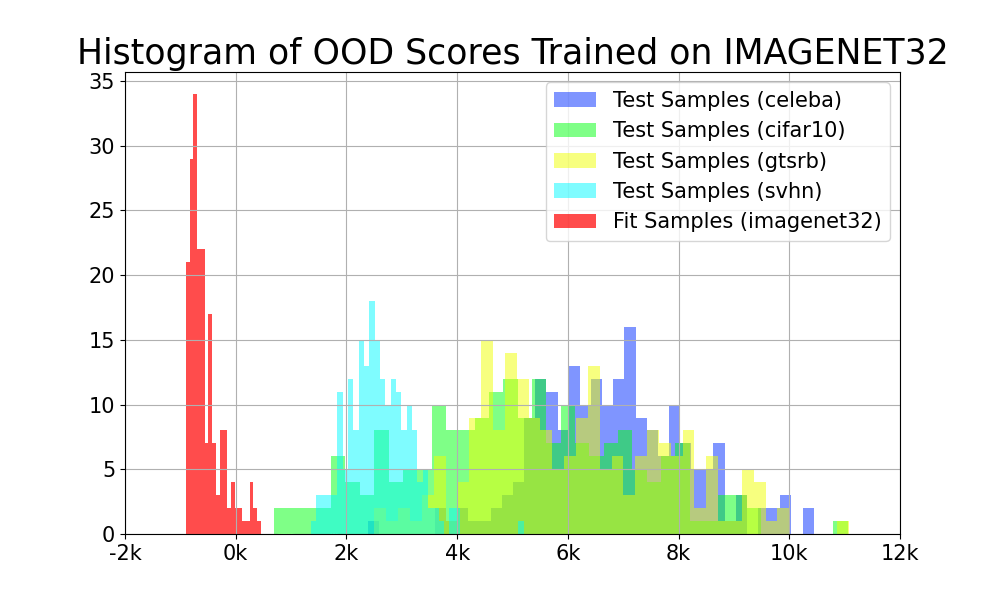}
    \end{subfigure}
    \hfill
    \begin{subfigure}[t]{0.49\textwidth}
        \centering
        \includegraphics[width=\textwidth]{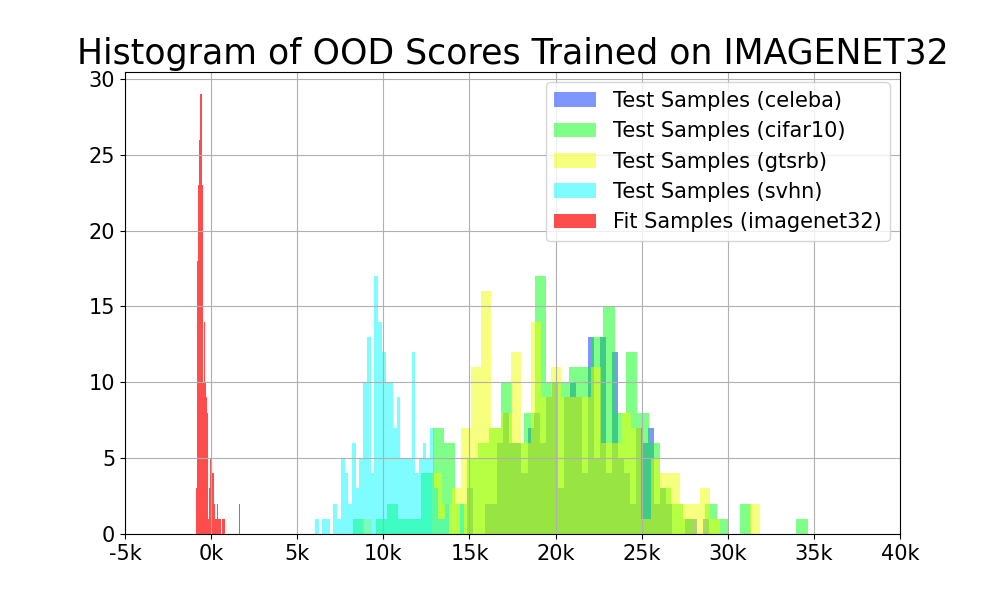}
    \end{subfigure}
    \caption{\textbf{\emph{Progressive widening of the gap in histograms of layer-wise gradient-based OOD scores with batch size of $5$.}}\\[0.5em]
    The figure illustrates how training on a complex ID dataset, such as ImageNet32, affects the gap between histograms of OOD scores for ID and OOD samples from (GTSRB, CIFAR-10, CelebA, and SVHN). Figure (left) represents the results after 10 epochs, while figure (right) shows the results after 250 epochs. Despite the increasing gap as training progresses, AUROC scores remain unchanged compared to a partially trained model, indicating that early training may suffice for OOD detection tasks. This widening gap, while reflecting improved separation, incurs higher computational costs. Additional experiments, including results for other batch sizes (e.g., $b=1$) and ID datasets, are discussed in \S~\ref{sec:Additional_Results}.}
    \label{fig:layerwise_scores_complex_comparison}
\end{figure}

\noindent Extracting the OOD score as explained in \S~\ref{sec:Methodology}, two major phenomena can be observed: (1) training the model on a complex dataset (e.g., ImageNet32) and (2) training on a less complex dataset (e.g., SVHN). In the first case, as the model is trained to achieve lower bits per dimension (BPD) and potentially better image quality, the gap between the ID and OOD histograms of OOD scores becomes increasingly wider (see Fig. \ref{fig:layerwise_scores_complex_comparison}). However, this widening gap does not significantly affect the AUROC results. Therefore, a minimal yet sufficient gap, achievable during earlier training epochs, is enough to yield AUROC results comparable to those of a fully converged model (see Tables \ref{tab:auc_results_three_channel} and \ref{tab:auc_results_one_channel}). 

\begin{figure}[ht!]
    \centering
    \begin{subfigure}[t]{0.49\textwidth}
        \centering
        \includegraphics[width=\textwidth]{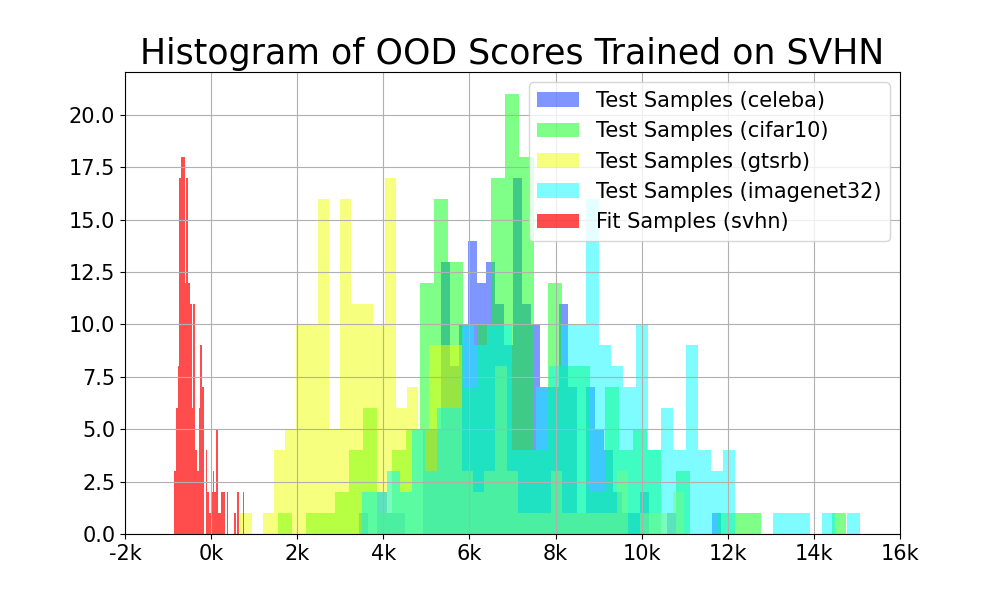}
    \end{subfigure}
    \hfill
    \begin{subfigure}[t]{0.49\textwidth}
        \centering
        \includegraphics[width=\textwidth]{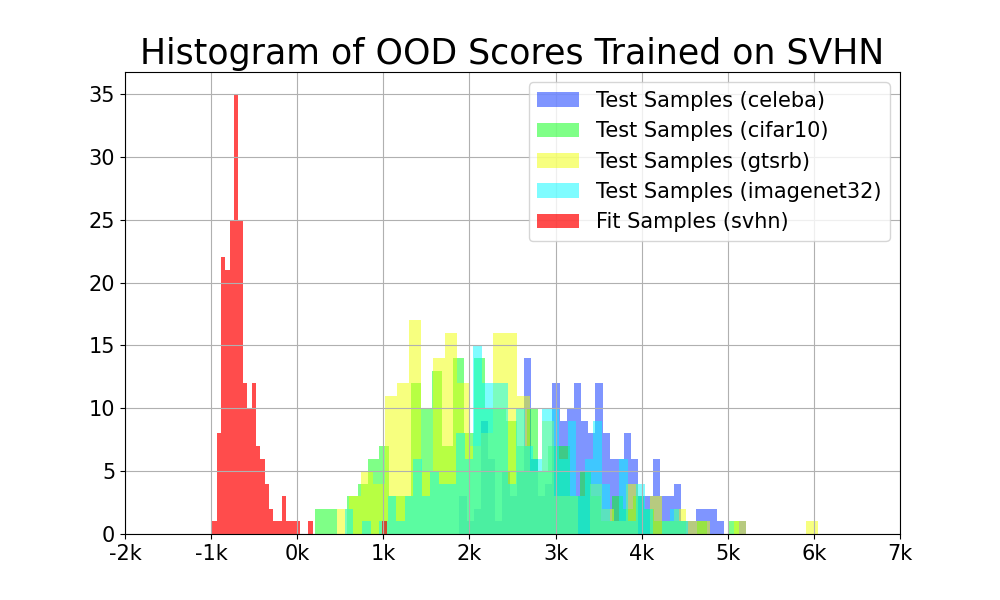}
    \end{subfigure}
    \caption{\textbf{\emph{Transition from gap to overlap in histograms of layer-wise gradient-based OOD scores using batch size of $5$ as training progresses.}}\\[0.5em]This phenomenon occurs when the ID dataset is less complex compared to the OOD samples, resulting in a decline in OOD detection performance with a fully converged model that generates higher-quality images. The GLOW model was trained on the ID dataset SVHN for: (left) one epoch and (right) 250 epochs, and tested against four OOD datasets: GTSRB, CIFAR-10, CelebA, and ImageNet32. As training progresses, the distinct separation between ID and OOD data distributions deteriorates. Additional results with varying batch sizes (e.g., $b=1$ and $b=5$) and other IDs are detailed in \S~\ref{sec:Additional_Results}.}
    \label{fig:layerwise_scores_simple_comparison}
\end{figure}

\noindent In the second case, when the model is trained on a simpler dataset, as it reduces the BPD, its ability to detect OOD samples diminishes. The gap observed in the earlier epochs transitions into overlap in the fully trained model or an exiting overlap becomes more pronounced (refer to Table \ref{tab:auc_results_one_channel}). This suggests that OOD detection primarily operates on low-level image features (e.g., textures and curves) rather than high-level features (e.g., objects in images) (see Fig. \ref{fig:layerwise_scores_simple_comparison}).

\noindent These phenomena are consistent across different combinations of ID and OOD datasets for both three-channel and one-channel colour images (additional comparisons are provided in \S~\ref{sec:Additional_Results}). Notably, for certain datasets, such as SVHN and Omniglot, effective OOD detection can be achieved using only the first epoch, leading to better AUROC results. It is important to note that if the size of the training dataset is sufficiently large relative to the dataset's complexity, allowing the model to adequately learn the ID samples, a gap will emerge between the histograms of OOD scores. This indicates a clear distinction between ID and OOD samples, as seen in datasets like Imagenet32, Celeba, SVHN, and Omniglot. However, synthetically increasing the dataset size—such as through data augmentation, adding generated samples to the training set, or utilizing mix-up/cutout techniques—does not significantly boost AUROC scores to nearly perfect levels for distinguishing singleton OOD samples, as demonstrated with CIFAR-10 and GTSRB. Therefore, their results are not reported.

\begin{figure}[ht!]
    \centering
    \begin{subfigure}[t]{0.49\textwidth}
        \centering
        \includegraphics[width=\textwidth]{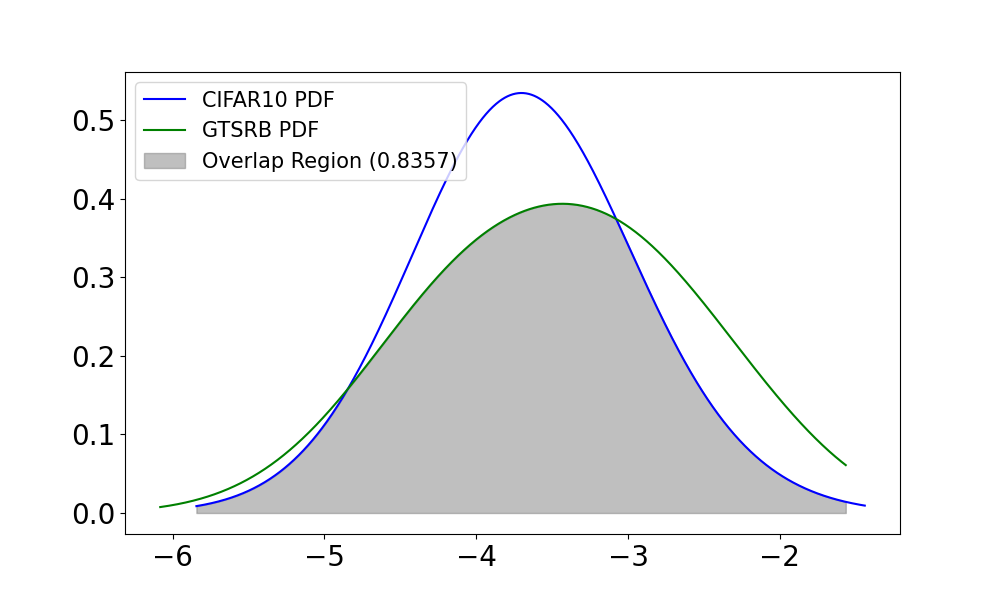}
    \end{subfigure}
    \hfill
    \begin{subfigure}[t]{0.49\textwidth}
        \centering
        \includegraphics[width=\textwidth]{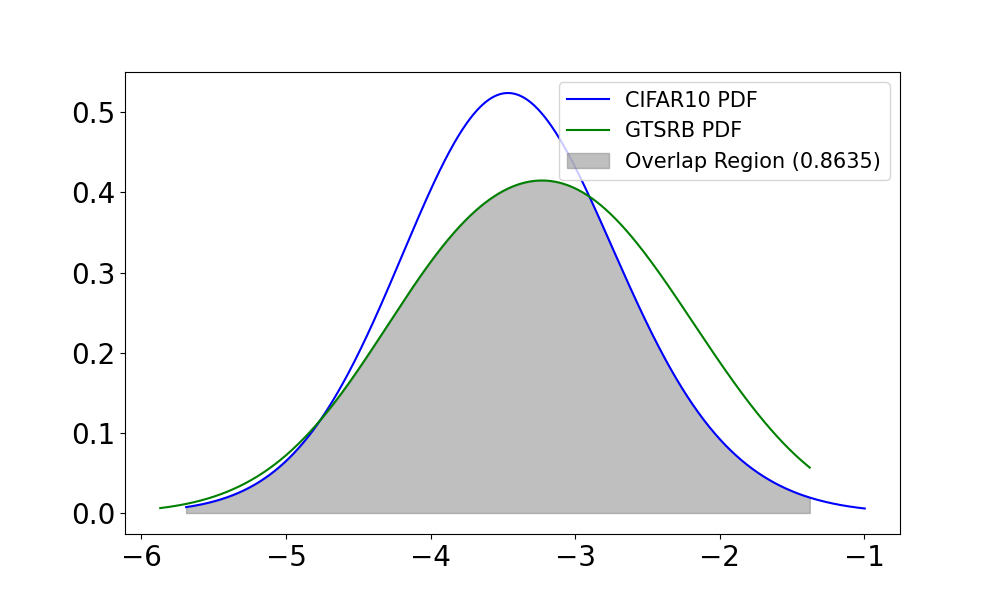}
    \end{subfigure}
    \caption{\textbf{\emph{Overlap Coefficient (OVL)}}\\[0.5em] Overlap area (\cite{walker2021newmeasure, weitzman1970overlap} ) between the PDFs of negative BIDs for an ID dataset (CIFAR-10) and an OOD dataset (GTSRB), evaluated using the Glow model trained on CIFAR-10 at different epochs. Figure (a) shows the OVL value of 0.8357 after 50 epochs, and (b) value of 0.8635 for the fully trained model (lower values are better). The increase in OVL reflects a larger overlap of distributions as training progresses, which correlates with a decline in AUC values of OOD detection: 0.6310 (50 epochs), and 0.5502 (fully trained), (higher values are better).}
    \label{fig:support_overlap}
\end{figure}

\noindent\\ Consider the case when NLL is used as the OOD score. Using the general concept of support overlap, it can be observed that as the model progresses in training, the support overlap between the ID and OOD distributions increases. This increasing overlap degrades the quality of OOD scores generated by statistics such as log-likelihood or its combinations (see Fig. \ref{fig:support_overlap}). Formally, when the supports of the in-distribution \(P\) and the out-distribution \(Q\) overlap (i.e., there exists \(x \in X\) such that both \(P(x) > 0\) and \(Q(x) > 0\)), a misestimated model \(P_\theta\), constructed based on the likelihood ratio, can lead to improved OOD detection performance compared to using the true distribution \(P\). Specifically, \(P_\theta(x)\) is defined as being proportional to the likelihood ratio of \(P(x)\) and \(Q(x)\), normalized by a constant \(C\) assuming integrability:

\[
P_\theta(x) = \frac{1}{C} \frac{P(x)}{Q(x)}, \quad \text{where } C = \int_X \frac{P(x)}{Q(x)} \, dx.
\]

\noindent Using this construction, the test statistic \(\phi_{P_\theta}\) becomes proportional to the likelihood ratio \(\frac{P(x)}{Q(x)}\). According to the Neyman-Pearson Lemma (\cite{neyman1933most}), the likelihood ratio test is uniformly most powerful for distinguishing \(P\) from \(Q\). As a result, OOD detection using \(P_\theta\) achieves a higher Area Under the Curve (AUC) than detection using the true distribution \(P\), expressed as:

\[
\text{Pr}(\phi_{P_\theta}(x) > \phi_{P_\theta}(y)) > \text{Pr}(\phi_P(x) > \phi_P(y)),
\]

\noindent where \(x \sim P\) and \(y \sim Q\). This improvement holds for any model \(P_\theta\) that makes values of \(\phi_{P_\theta}(x)\); \(x \sim P\) higher
relative to \(\phi_{P_\theta}(y)\); \(y \sim Q\) leading to more effective OOD detection.

\section{Experimental Results}
\label{sec:Experimental_Results}
\subsection{OOD Results}
\label{sec:OOD_Results}

In this section, the results of a partially trained versus fully trained model is demonstrated. To maintain comparability of results, GLOW (\cite{kingma2018glow}) as the generative model is employed and trained for 250 epochs on five three-channel datasets: \emph{ImageNet32, CIFAR-10, CelebA, GTSRB, and SVHN}, with all images resized to a uniform resolution of \( 32 \times 32 \) pixels
and in another experiment using four one-channel datasets \emph{MNIST, FashionMNIST, KMNIST and Omniglot} with a resolution of \( 28 \times 28 \), however, limited the selection of epochs to $(1st, 10th, 20th, 30th, 40th, 50th, 70th, 80th, 100th, 150th )$ as the nominee of partially training and the last epoch as fully training.

\noindent For each experiment, one dataset is designated as the ID dataset, while the remaining datasets in each family channel are treated as OOD. The training split of the ID dataset is utilized to train the model and the test split to fit Gaussian distributions to the logarithmic  trained parameters. From the test splits of all datasets, 1,000 samples are randomly selected to compute OOD scores and generate AUROC results, as summarized in Tables~\ref{tab:auc_results_three_channel} and ~\ref{tab:auc_results_one_channel}.

\noindent The effectiveness of OOD detection is evaluated based on the test's ability to correctly classify OOD samples while minimizing the misclassification of ID samples as OOD. This performance is calculated through a Receiver Operating Characteristic (ROC) curve, which checks the True Positive Rate (TPR)—the proportion of correctly identified OOD samples—against the False Positive Rate (FPR)—the proportion of misclassified ID samples. The ROC curve illustrates the trade-off between sensitivity (TPR) and specificity (\( 1 - \text{FPR} \)) across various threshold settings.

\noindent The Area Under the ROC Curve (AUROC) provides a single metric to quantify the overall performance of the OOD detection method. A higher AUROC value, closer to 1, reflects better performance, indicating the test's ability to effectively distinguish OOD samples from ID samples. As a summary statistic, the AUROC captures the balance between correctly detecting OOD samples and avoiding the misclassification of ID samples.

\begin{table}[ht!]
\centering
\begin{tabular}{lll|cccc}
\toprule
\textbf{(Train ↓)}\textbf{(Test →)}&&\textbf{Epochs} & \textbf{MNIST} & \textbf{FashionMNIST} & \textbf{KMNIST} & \textbf{Omniglot} \\
\midrule
\multirow{6}{*}{\textbf{MNIST}} 
    & & $80^{th}$      && \textbf{0.9666} & \textbf{0.9522} &\textbf{1.0000} \\
    &$(b=1)$ &fully trained  &-& 0.9653 & 0.9511 & \textbf{1.0000} \\
\cmidrule{2-7}
    && $80^{th}$      && \textbf{0.9995} & 0.9997 &\textbf{1.0000} \\
    & $(b=5)$ &fully trained  &-& \textbf{0.9995} & \textbf{0.9998}&\textbf{1.0000} \\

\midrule
\midrule
\multirow{6}{*}{\textbf{FashionMNIST}}   
    & &$70^{th}$       & \textbf{0.9547}   &&\textbf{0.9025}&\textbf{1.0000}  \\
    &$(b=1)$ &fully trained    & 0.9546   &-& 0.9020 & \textbf{1.0000} \\
\cmidrule{2-7}
    &  &$50^{th}$       & \textbf{0.9991}&& \textbf{0.9936}&\textbf{1.0000}  \\
    &$(b=5)$ &fully trained& 0.9985&-& 0.9935 & \textbf{1.0000}\\
\midrule
\midrule
\multirow{6}{*}{\textbf{KMNIST}}
    & &$30^{th}$      & \textbf{0.4715}   & \textbf{0.7315} && \textbf{0.9983}\\
    & $(b=1)$&fully trained   & 0.4661   & 0.6737 &-& 0.9960\\
\cmidrule{2-7}
    &&$20^{th}$      & \textbf{0.6595}   & \textbf{0.9590} &&\textbf{1.0000}\\
    & $(b=5)$ &fully trained   & 0.6290   &0.9499 &-& \textbf{1.0000}\\
\midrule
\midrule
\multirow{6}{*}{\textbf{Omniglot}}    
    & &$1^{st}$&  \textbf{1.0000}&\textbf{1.0000} &\textbf{1.0000}& \\
    &$(b=1)$ &fully trained&\textbf{1.0000}&\textbf{1.0000} &\textbf{1.0000}&-\\
\cmidrule{2-7}
    & &$1^{st}$&  \textbf{1.0000}&\textbf{1.0000}&\textbf{1.0000}&\\
    &$(b=5)$ &fully trained& \textbf{1.0000}&\textbf{1.0000}&\textbf{1.0000}&-\\
\bottomrule
\end{tabular}
\vspace{0.5cm}

\caption{\textbf{\emph{AUROC Results for One-Channel colour Datasets}}\\[0.5em]
This table presents the AUROC results for various ID and OOD dataset pairs (\emph{MNIST, FashionMNIST, KMNIST, Omniglot}) trained with GLOW and evaluated at two training stages: partial training (\emph{e.g., 1st, 20th, 30th, 50th, 70th, and 80th epochs}) and fully trained models. Results are reported for single OOD detection (\(b=1\)) and batch size \(b=5\) using the OOD scores defined in Equation~\ref{eq:ood_score_final}. For most dataset pairs, partial training—depending on the dataset—outperforms fully trained models in OOD detection, even as the fully trained models produce higher-quality images. Bolded values highlight the highest AUROC scores, indicating the most effective training stage for distinguishing OOD data. Generated samples from each model can be found in \S~\ref{sec:Generated_Samples}.}
\label{tab:auc_results_one_channel}
\end{table}

\begin{table}[ht!]
\centering
\begin{tabular}{lll|ccccc}
\toprule
\textbf{(Train ↓)}\textbf{(Test →)} & &\textbf{Epochs} & \textbf{SVHN} & \textbf{CelebA} & \textbf{CIFAR-10} & \textbf{GTSRB} & \textbf{ImageNet32} \\
\midrule
\multirow{6}{*}{\textbf{SVHN}}
    &&$10^{th}$&& \textbf{0.9852} & \textbf{0.9765} & \textbf{0.9561} & \textbf{0.9833} \\
    &$(b=1)$&fully trained&-& 0.9377 & 0.9243 & 0.9215 & 0.9091 \\
\cmidrule{2-8}
    & &$1^{st}$& & \textbf{1.0000}&\textbf{1.0000} & \textbf{0.9998}& \textbf{1.0000} \\
    &$(b=5)$&fully trained&-& \textbf{1.0000} & 0.9994 & 0.9996 & 0.9999 \\
\midrule
\midrule
\multirow{6}{*}{\textbf{CelebA}}   
    &&$50^{th}$& \textbf{0.9896}&& \textbf{0.9799} & \textbf{0.9755} & \textbf{0.9865} \\
    &$(b=1)$&fully trained& 0.9876 &-& 0.9797 & 0.9753 & 0.9790 \\
\cmidrule{2-8}
    &&$20^{th}$& \textbf{1.0000}&& \textbf{1.0000} & \textbf{1.0000}& \textbf{1.0000} \\
    &$(b=5)$&fully trained& \textbf{1.0000}&-& \textbf{1.0000} & \textbf{1.0000}& \textbf{1.0000} \\
\midrule
\midrule
\multirow{6}{*}{\textbf{CIFAR-10}}
    &&$100^{th}$& \textbf{0.8652}& \textbf{0.5410} && \textbf{0.7191}& 0.5893 \\
    &$(b=1)$&fully trained& 0.8185 & 0.4142 &-& 0.6798 & \textbf{0.7001} \\
\cmidrule{2-8}
    &&$50^{th}$& \textbf{0.9987}& \textbf{0.9377} && \textbf{0.9388} & 0.8247 \\
    &$(b=5)$&fully trained& 0.9984& 0.9140 &-& 0.9140 & \textbf{0.9480} \\
\midrule
\midrule
\multirow{6}{*}{\textbf{GTSRB}}  
    &&$150^{th}$& \textbf{0.9583}& \textbf{0.9249} & 0.9079 && \textbf{0.9509} \\
    &$(b=1)$&fully trained& 0.9579& 0.9177 & \textbf{0.9221} &-& 0.9499 \\
\cmidrule{2-8}
    &&$100^{th}$& \textbf{1.0000}& \textbf{0.9999} & 0.9990 & -& 0.9999 \\
    &$(b=5)$&fully trained& \textbf{1.0000}& \textbf{0.9999} & \textbf{0.9997} && \textbf{1.0000} \\
\midrule
\midrule
\multirow{6}{*}{\textbf{ImageNet32}} 
    &&$40^{th}$& 0.8949& \textbf{0.9999} & \textbf{0.9996} & \textbf{0.9999} & \\
    &$(b=1)$&fully trained& \textbf{0.9994}& 0.9998 & \textbf{0.9996} & 0.9998 &- \\
\cmidrule{2-8}
    &&$10^{th}$& \textbf{1.0000}& \textbf{1.0000} & \textbf{1.0000} & \textbf{1.0000} &\\
    &$(b=5)$&fully trained & \textbf{1.0000}& \textbf{1.0000} & \textbf{1.0000} & \textbf{1.0000}&-\\
\bottomrule
\end{tabular}
\vspace{0.5cm}

\caption{\textbf{\emph{Three-channel colour Dataset AUROC Results}}\\[0.5em]
The AUROC results for OOD detection across various three-channel datasets (\emph{SVHN}, \emph{CelebA}, \emph{CIFAR-10}, \emph{GTSRB}, and \emph{ImageNet32}) evaluated at different training epochs (e.g., 1st, 10th, 20th, 40th, 50th, 100th, 150th and fully trained). The OOD detection scores, calculated as $S_{\bm{\theta}^{(l)}}(\bm{x}_{b=\{1,5\}}) = \log \big\{ \big\| \nabla_{\bm{\theta}^{(l)}} \big( \sum_b \ell(\bm{x}_b) \big) \big\|_2^2 \big\}$, are provided for both batch sizes of 1 and 5. Each row corresponds to a specific in-distribution (ID) dataset (Train ↓) evaluated against various out-of-distribution (OOD) datasets (Test →). Bolded values highlight the highest AUROC scores achieved for a given dataset pair across the training epochs. The results show that in many cases, the Glow model achieves near-optimal or optimal OOD detection performance after partial training, often surpassing or equalling the performance of the fully trained model. Notably, the epochs selected for each dataset may not necessarily be the lowest but are instead sampled from every tenth epoch, demonstrating that partial training can outperform or match fully converged models. These findings emphasize the diminishing returns of extended training for OOD detection tasks in many scenarios. See generated samples \ref{sec:Generated_Samples} and visualization of results in \ref{sec:Additional_Results}.}
\label{tab:auc_results_three_channel}
\end{table}

\section{Conclusion}
\label{sec:Conclusion}
This study demonstrates that using layer-wise changes in parameter norms with respect to individual data points, partially trained models outperform fully trained models in OOD detection tasks. The results suggest that the models excel in identifying low-level features such as dominant colors, textures, and structures, which are more prominent in earlier training stages. However, this work has two key limitations. First, future experiments should explore other DGMs to validate the findings across a broader range of architectures. Second, while this study provides an extensive empirical benchmark for DGMs, OOD, and ID datasets, future research should expand beyond image datasets to include other data modalities, such as text, and evaluate performance on large language models.

\section*{Acknowledgments}
\label{sec:Acknowledgments}
The authors would like to thank Felix Draxler, Armand Rousselot and members of CVL for their guidance and insightful discussions throughout this work.

\printbibliography

\appendix
\section{Additional Results}
\label{sec:Additional_Results}

This section provides a comprehensive overview of how partial training can be beneficial for OOD detection. It presents visualizations of histograms of AUROC scores (from Tables \ref{tab:auc_results_one_channel} and \ref{tab:auc_results_three_channel}) based on 1000 random samples from various datasets, using the OOD score defined in equation \ref{eq:ood_score_final}. These histograms demonstrate that partially trained models can outperform, or at least match, fully trained models in OOD detection tasks. This challenges the assumption that longer training always leads to better OOD detection performance.

\begin{figure}[ht!]
    \centering
    \begin{subfigure}[t]{0.49\textwidth}
        \centering
        \framebox[\textwidth]{\includegraphics[width=\textwidth]{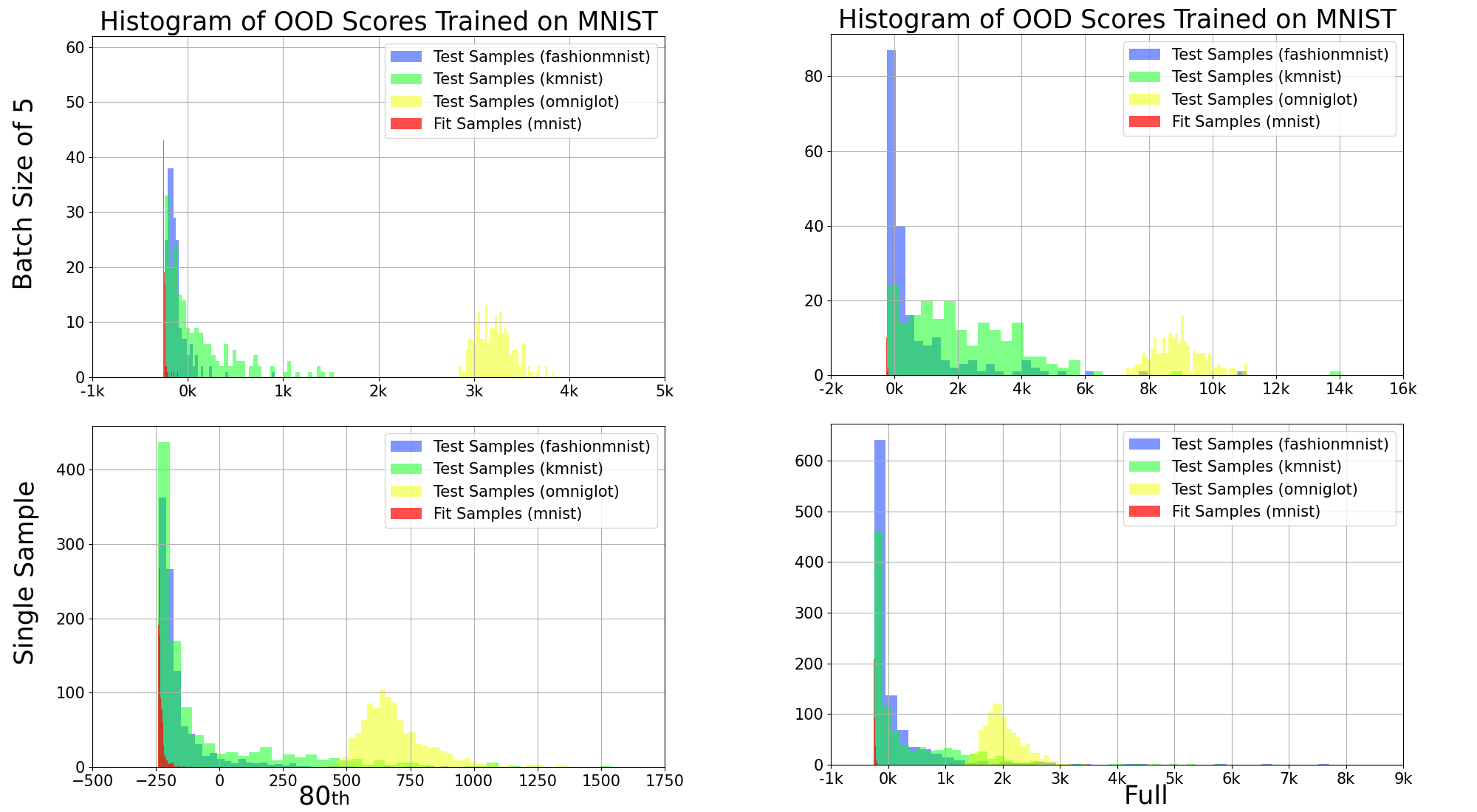}}
    \end{subfigure}
    \hfill
    \begin{subfigure}[t]{0.49\textwidth}
        \centering
        \framebox[\textwidth]{\includegraphics[width=\textwidth]{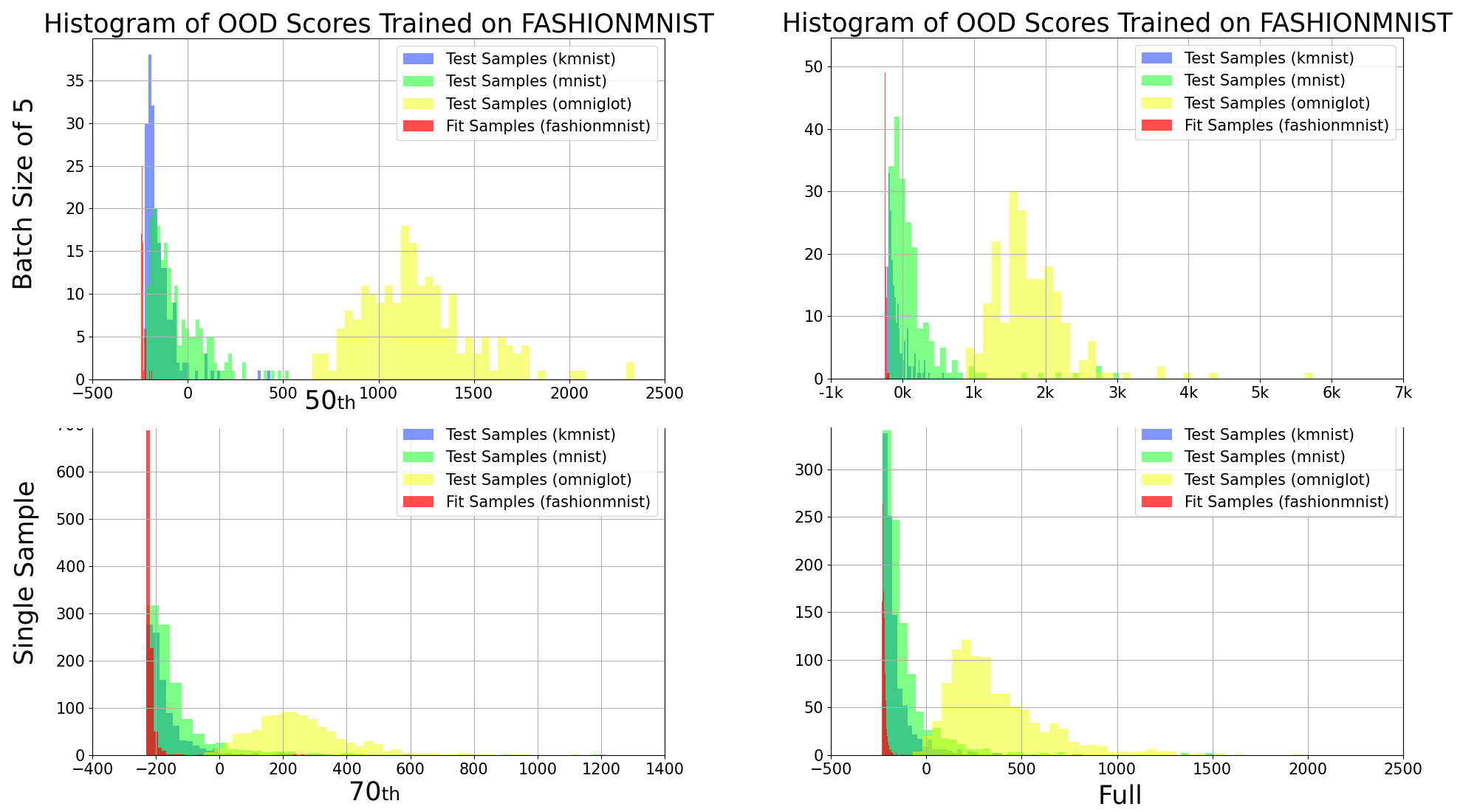}}
    \end{subfigure}
    \hfill
    \begin{subfigure}[t]{0.49\textwidth}
        \centering
       \framebox[\textwidth]{\includegraphics[width=\textwidth]{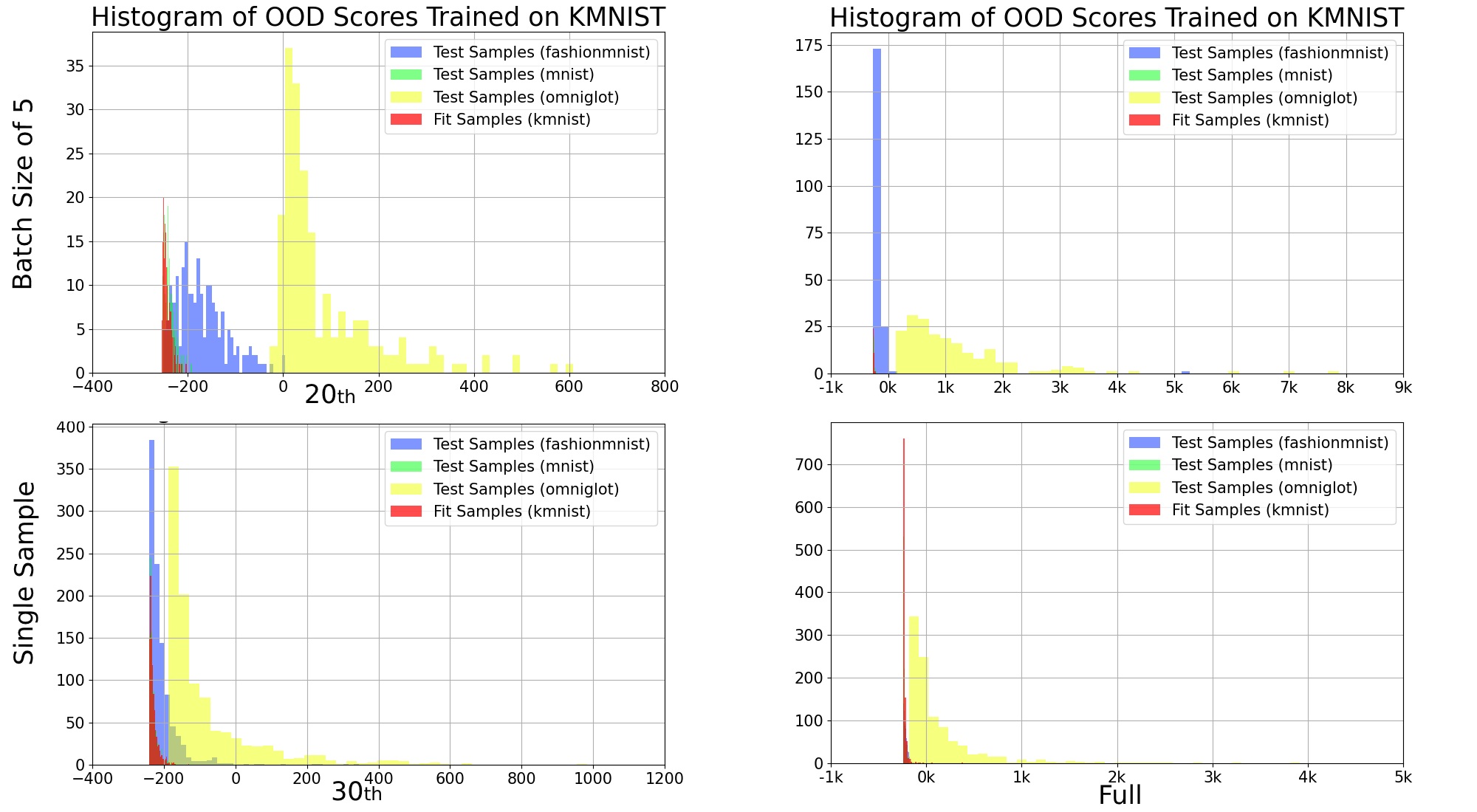}}
    \end{subfigure}
    \hfill
    \begin{subfigure}[t]{0.49\textwidth}
        \centering
        \framebox[\textwidth]{\includegraphics[width=\textwidth]{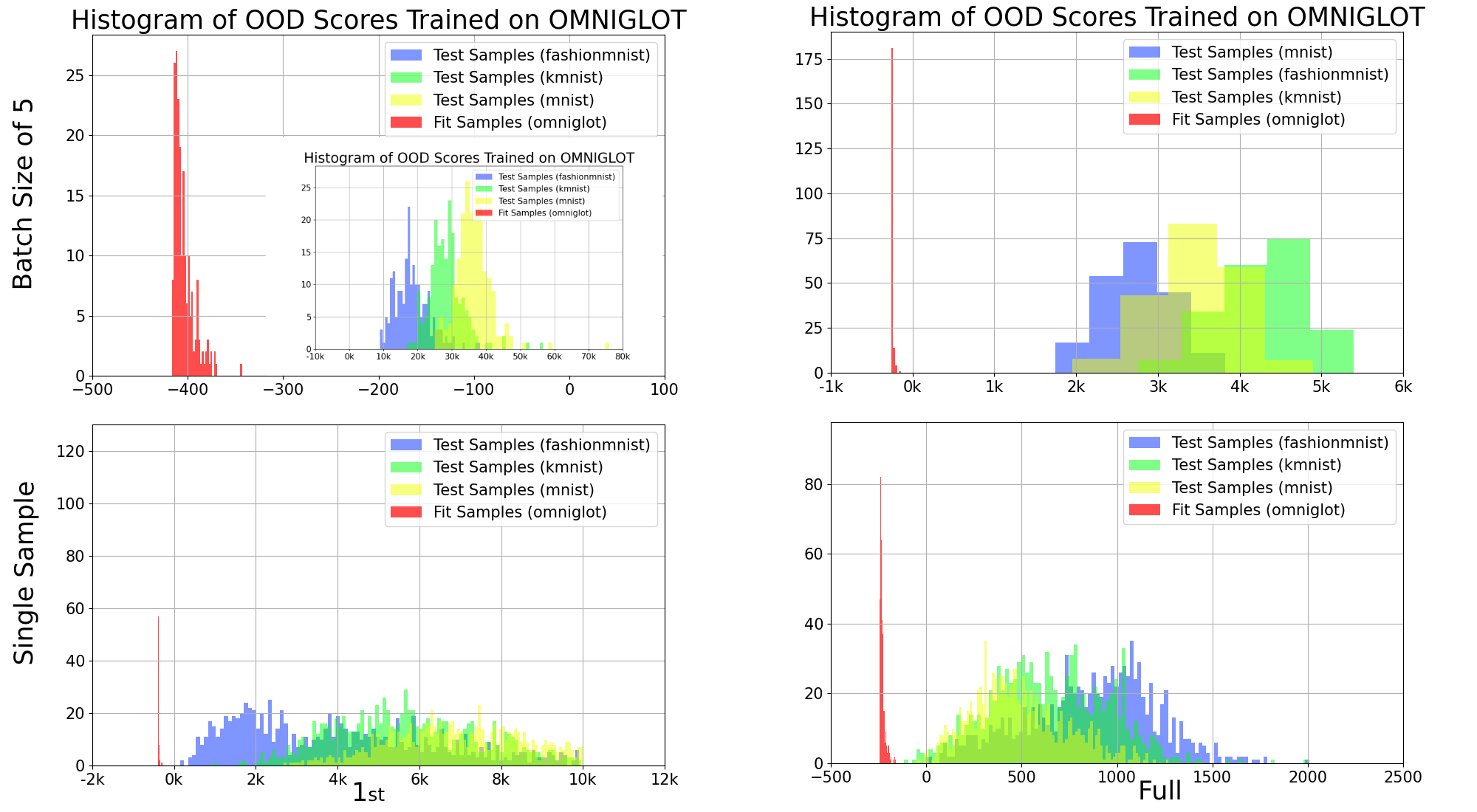}}
    \end{subfigure}
        \caption{\textbf{\emph{Histograms of OOD Scores for One-Channel colour Datasets}}\\[0.5em]
    The OOD scores, defined in Equation \ref{eq:ood_score_final}, are visualized here to provide insight into the performance of the models on one-channel datasets. These scores are linked to the AUROC scores presented in Table \ref{tab:auc_results_one_channel}, emphasizing the diminishing importance of fully trained models for OOD detection. Each box in the histograms represents the OOD scores when a dataset is treated as ID and evaluated against three different OOD datasets. The first row shows the results for OOD detection with a batch size of five, while the second row presents results for single-sample OOD detection. The first columns correspond to partially trained models at various epochs, while the second columns represent fully trained models at their final epochs.\\ The gap in the histograms of \emph{Omniglot} for the first epoch and a batch size of 5 is so wide that two different scales have been used to illustrate the performance effectively. In all other datasets, the overlap between histograms increases as training progresses.\\ When viewed alongside the generated samples in Figure \ref{fig:generated_samples} and the numerical results in Table \ref{tab:auc_results_one_channel}, this comparison demonstrates that partially trained models can achieve comparable or even superior OOD detection performance compared to fully trained ones.}
    \label{fig:OOD_hist_one_channel}
\end{figure}

\begin{figure}[ht!]
    \centering
    \begin{subfigure}[t]{0.49\textwidth}
        \centering
        \framebox[\textwidth]{\includegraphics[width=\textwidth]{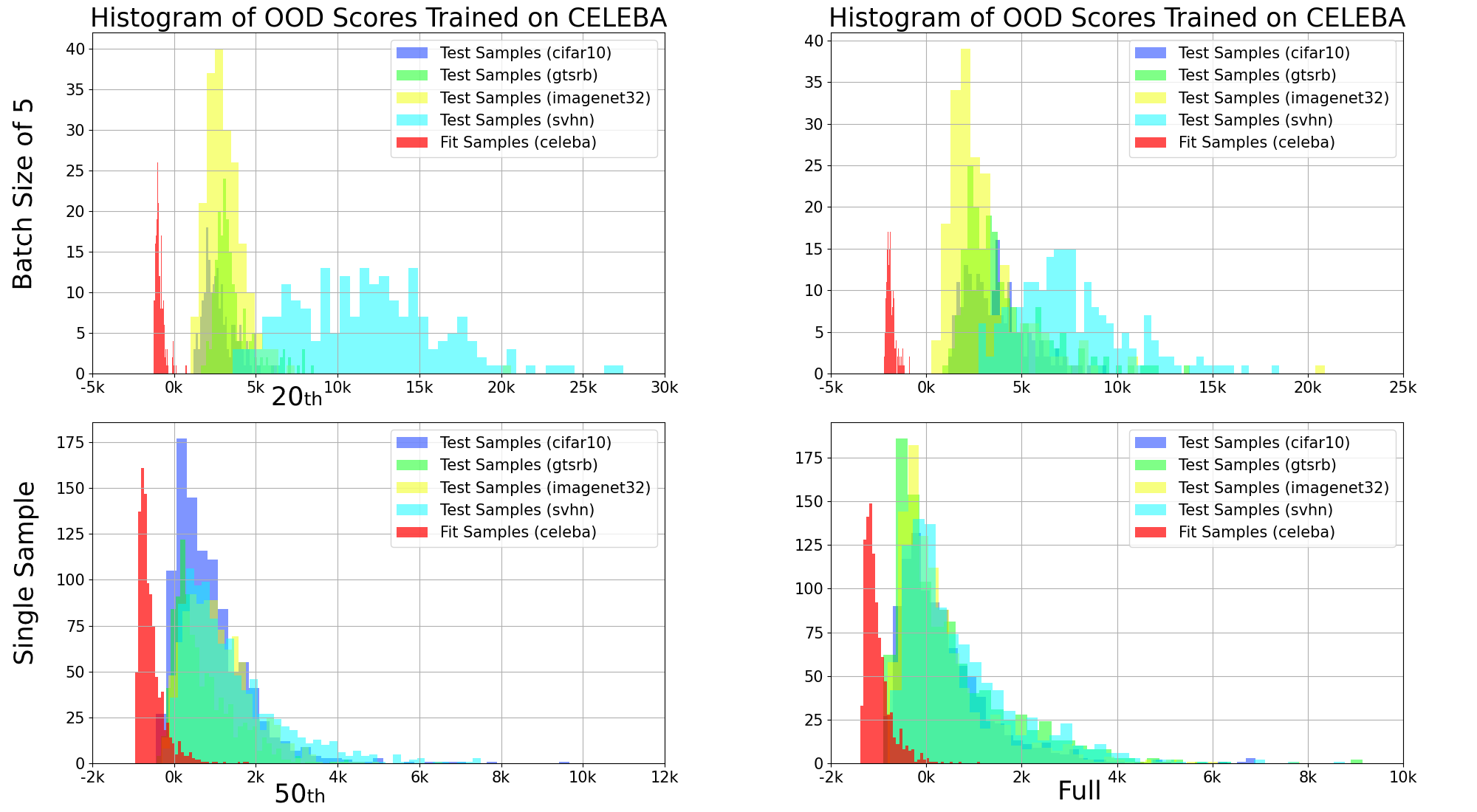}}
    \end{subfigure}
    \hfill
    \begin{subfigure}[t]{0.49\textwidth}
        \centering
        \framebox[\textwidth]{\includegraphics[width=\textwidth]{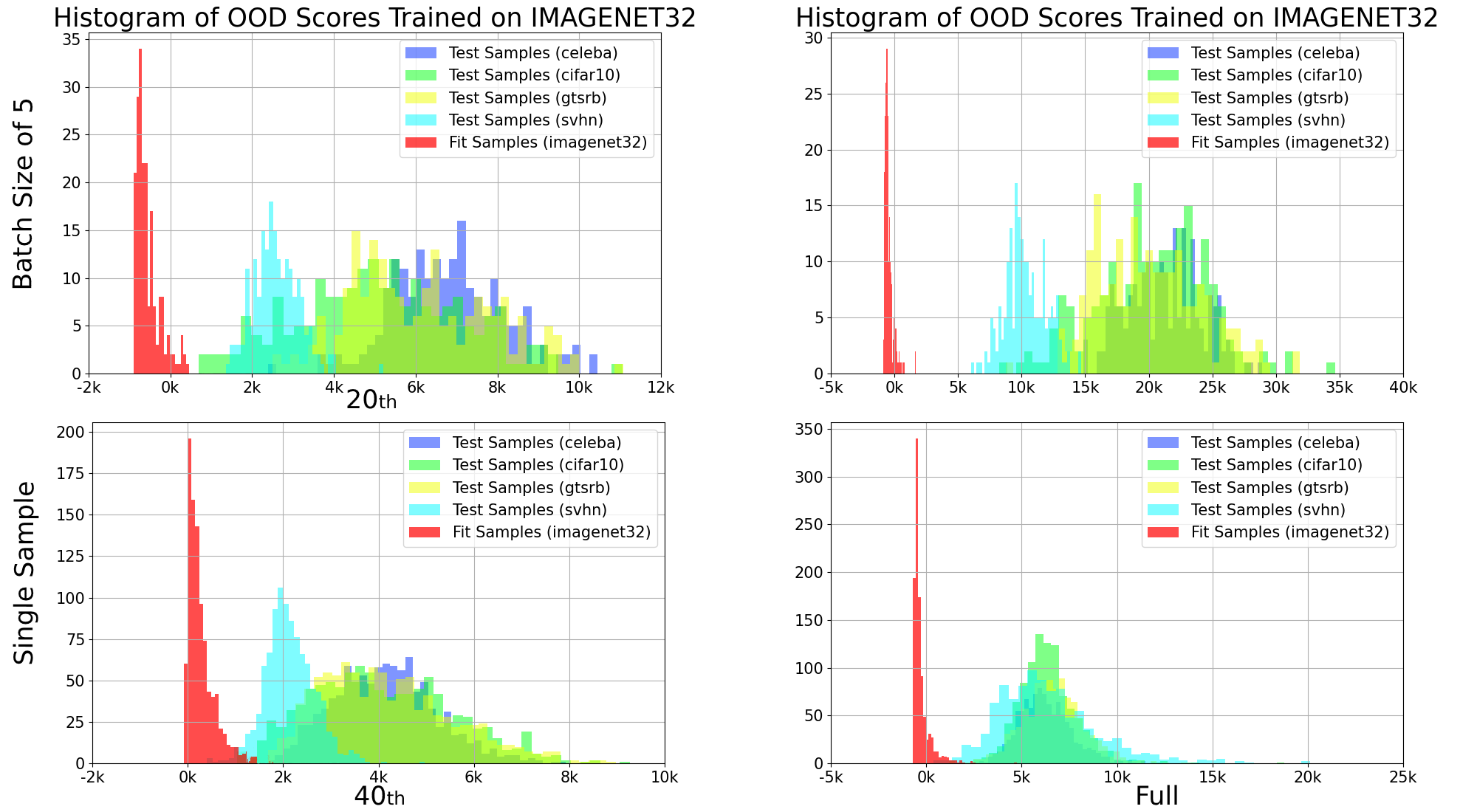}}
    \end{subfigure}
    \hfill
    \begin{subfigure}[t]{0.49\textwidth}
        \centering
        \framebox[\textwidth]{\includegraphics[width=\textwidth]{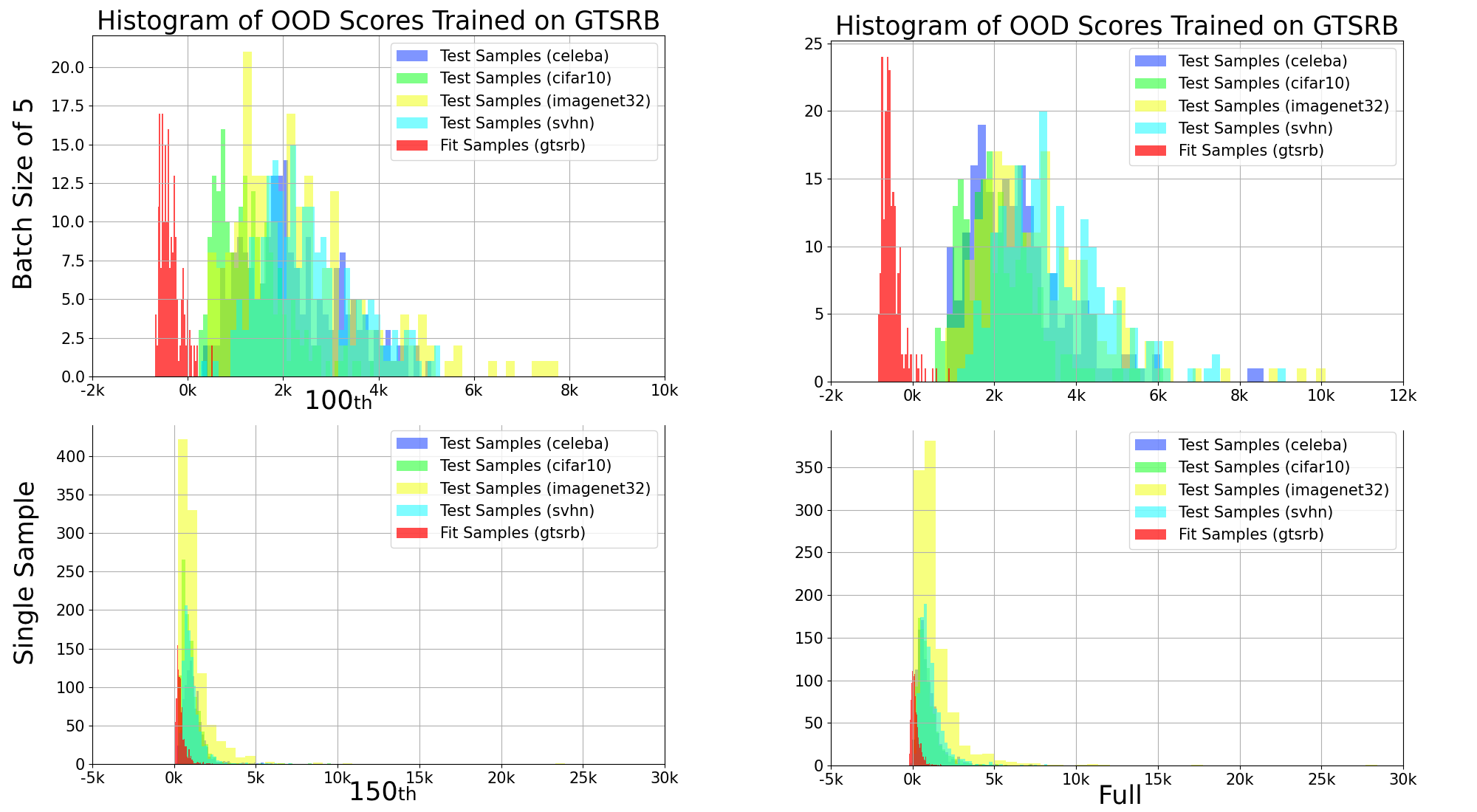}}
    \end{subfigure}
        \hfill
    \begin{subfigure}[t]{0.49\textwidth}
        \centering
        \framebox[\textwidth]{\includegraphics[width=\textwidth]{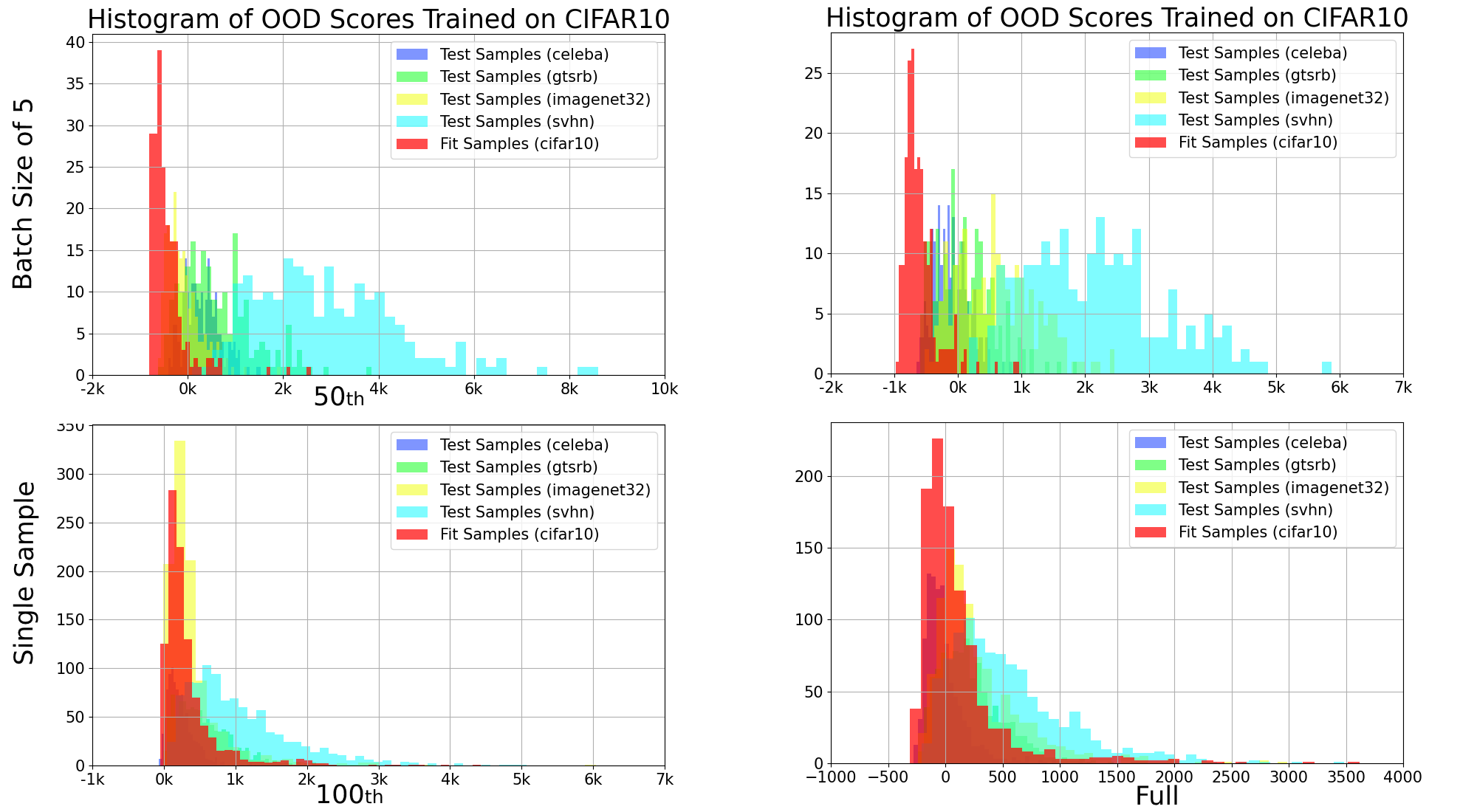}}
    \end{subfigure}
        \hfill
    \begin{subfigure}[t]{0.49\textwidth}
        \centering
        \framebox[\textwidth]{\includegraphics[width=\textwidth]{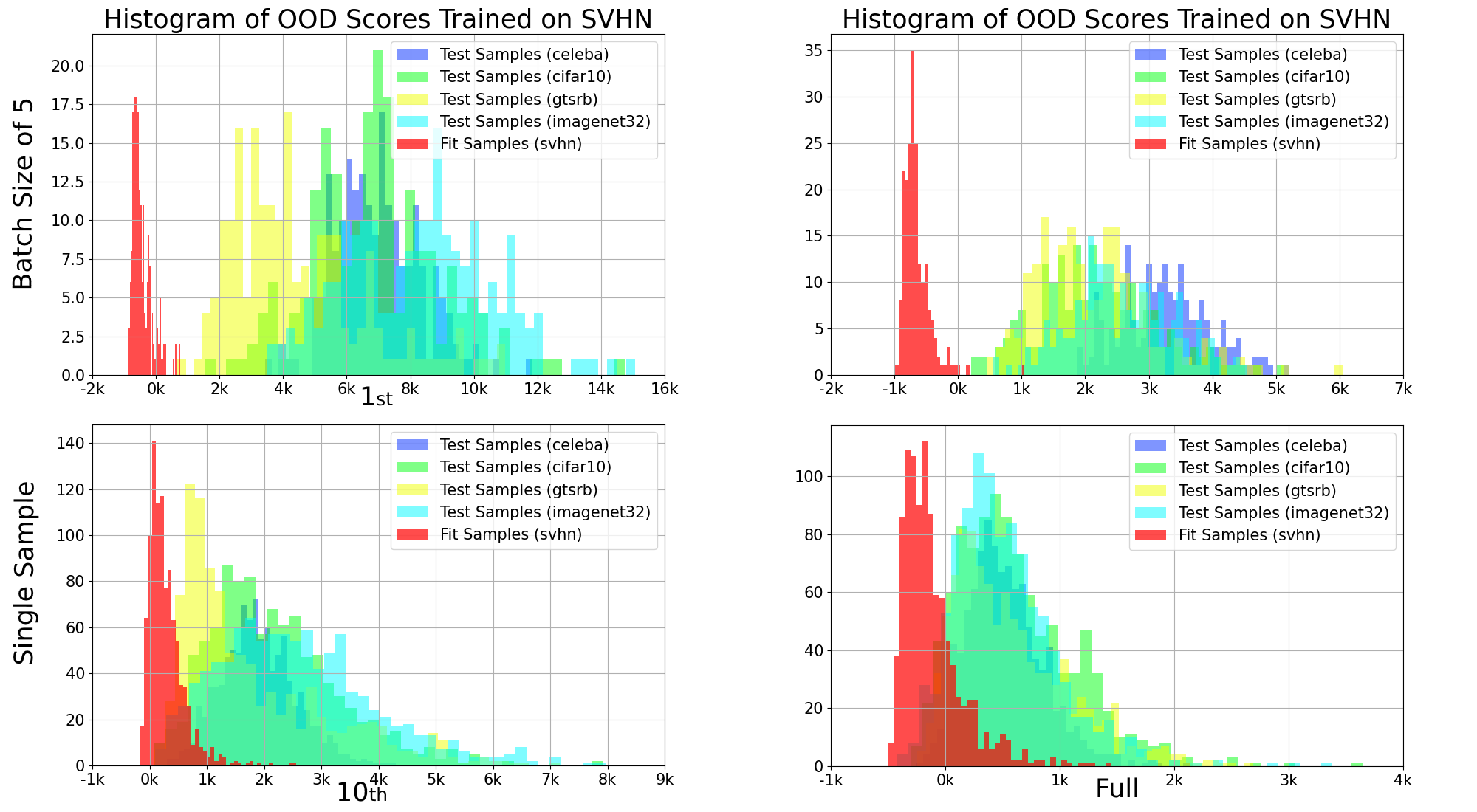}}
    \end{subfigure}

    \caption{\textbf{\emph{Histograms of OOD Scores for Three-Channel colour Datasets}}\\[0.5em] This figure presents the OOD scores for five three-channel datasets (CelebA, CIFAR-10, GTSRB, ImageNet32, and SVHN) at different stages of model training. The histograms illustrate the distribution of OOD scores when each dataset is treated as ID and evaluated against other datasets as OOD. The first row displays OOD detection results using a batch size of five, while the second row visualizes results for single-sample OOD detection. The first column corresponds to partially trained models at various epochs, whereas the second column shows results for fully trained models at the final epoch.\\
    Considering the first rows in boxes, for OOD detection using GoF, the gap between histograms increases with training progress for CelebA and ImageNet32. Conversely, in GTSRB and SVHN, the gap transitions into overlap, and for CIFAR-10, the overlap becomes more pronounced as training progresses. In single-sample OOD detection, the existing overlap becomes even more significant in fully trained models compared to partially trained ones.\\
    To better analyze the effectiveness of immature models, these histograms should be viewed alongside the generated sample images in Figure \ref{fig:generated_samples} and the AUROC results in Table \ref{tab:auc_results_three_channel}.}
    
    \label{fig:OOD_hist_three_channel}

\end{figure}

\clearpage
\section{Generated Samples}
\label{sec:Generated_Samples}
\begin{figure}[ht!]
    \centering
    \begin{subfigure}[t]{0.39\textwidth}
        \centering
        \includegraphics[width=\textwidth]{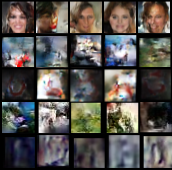}
    \end{subfigure}
    \begin{subfigure}[t]{0.39\textwidth}
        \centering
        \includegraphics[width=\textwidth]{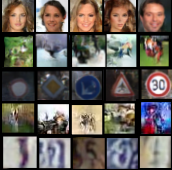}
    \end{subfigure}
    \begin{subfigure}[t]{0.39\textwidth}
        \centering
        \includegraphics[width=\textwidth]{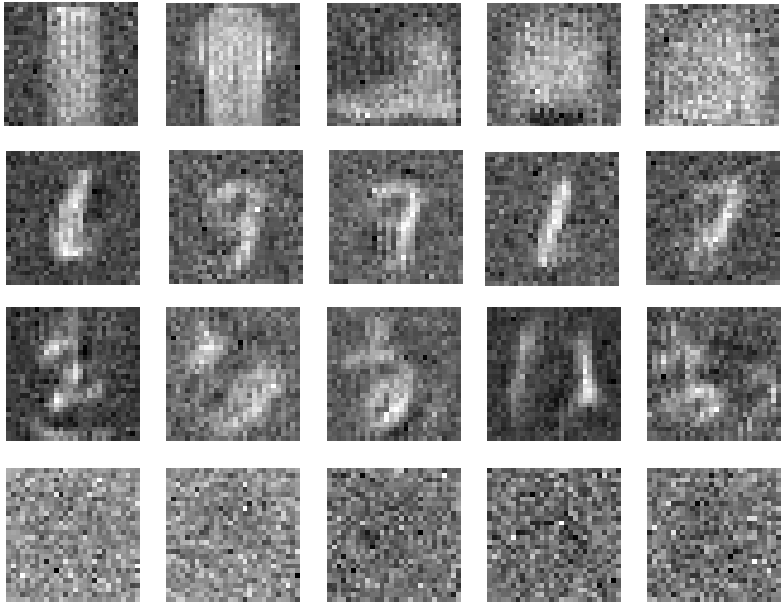}
    \end{subfigure}
    \begin{subfigure}[t]{0.39\textwidth}
        \centering
        \includegraphics[width=\textwidth]{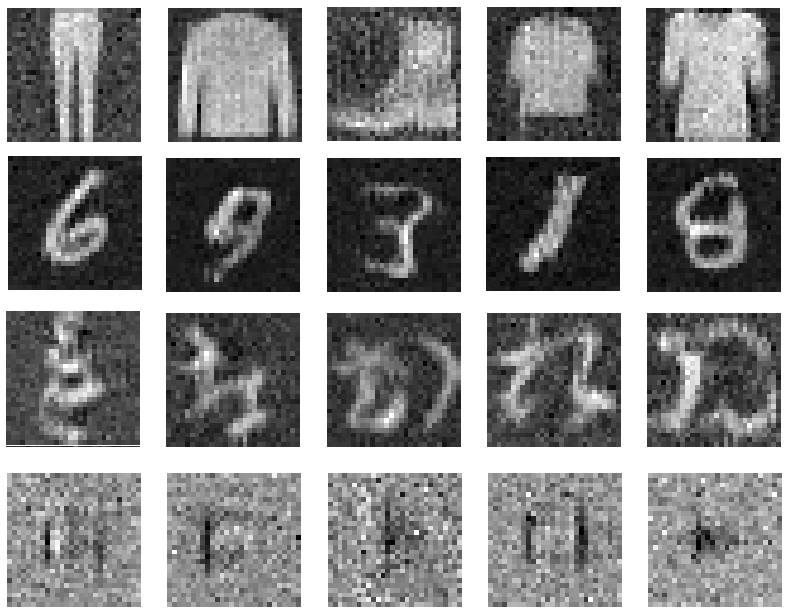}
    \end{subfigure}
\caption{\textbf{\emph{Generated Samples at Different Training Stages of the GLOW Model on Three-Channel and One-Channel Datasets}}\\[0.5em]
Samples generated from five three-channel datasets (CelebA, CIFAR-10, GTSRB, ImageNet32, and SVHN) and four one-channel datasets (FashionMNIST, MNIST, KMNIST, and Omniglot) at two different stages of training: partially trained (left) and fully trained (right). The samples were generated using a temperature parameter set to 1 across all datasets. As highlighted in Tables \ref{tab:auc_results_three_channel} and \ref{tab:auc_results_one_channel}, partially trained models frequently achieve higher or comparable AUROC scores in OOD detection tasks compared to fully trained models. The partially trained epochs correspond to intermediate training stages shown in the tables, while the fully trained models are evaluated at the 250th epoch for all three-channel datasets, the 200th epoch for FashionMNIST, MNIST, and KMNIST, and the 100th epoch for Omniglot. Notably, for Omniglot, effective OOD detection is evident even at the first epoch, negating the necessity for extended training.}
    \label{fig:generated_samples}
\end{figure}

Exploring the generated samples from the GLOW model trained on one-channel and three-channel datasets at different training stages. Figure \ref{fig:generated_samples} displays the generated samples for both types of datasets, with each set representing two stages: partially trained (left) and fully trained (right). The images for the three-channel datasets (CelebA, CIFAR-10, GTSRB, ImageNet32, and SVHN) are of resolution $32 \times 32$ pixels, while the one-channel datasets (FashionMNIST, MNIST, KMNIST, and Omniglot) have a resolution of $28 \times 28$ pixels. As training progresses, the quality of the generated samples improves, particularly for more complex datasets. Despite this improvement in sample quality, partially trained models often yield higher or equivalent AUROC scores in OOD detection tasks compared to fully trained models, as shown in the corresponding tables. This counter-intuitive relationship highlights a diminishing return in OOD detection performance as the models become fully converged. The intermediate epochs for the sample generation align with those presented in the respective tables, illustrating that partially trained models can often outperform fully trained models in the OOD detection task, even though their sample quality has not reached its peak.

\clearpage
\section{NLL as OOD Score}
\label{sec:NLL_as_OOD}
\begin{figure}[ht!]
    \centering
    \includegraphics[width=\textwidth]{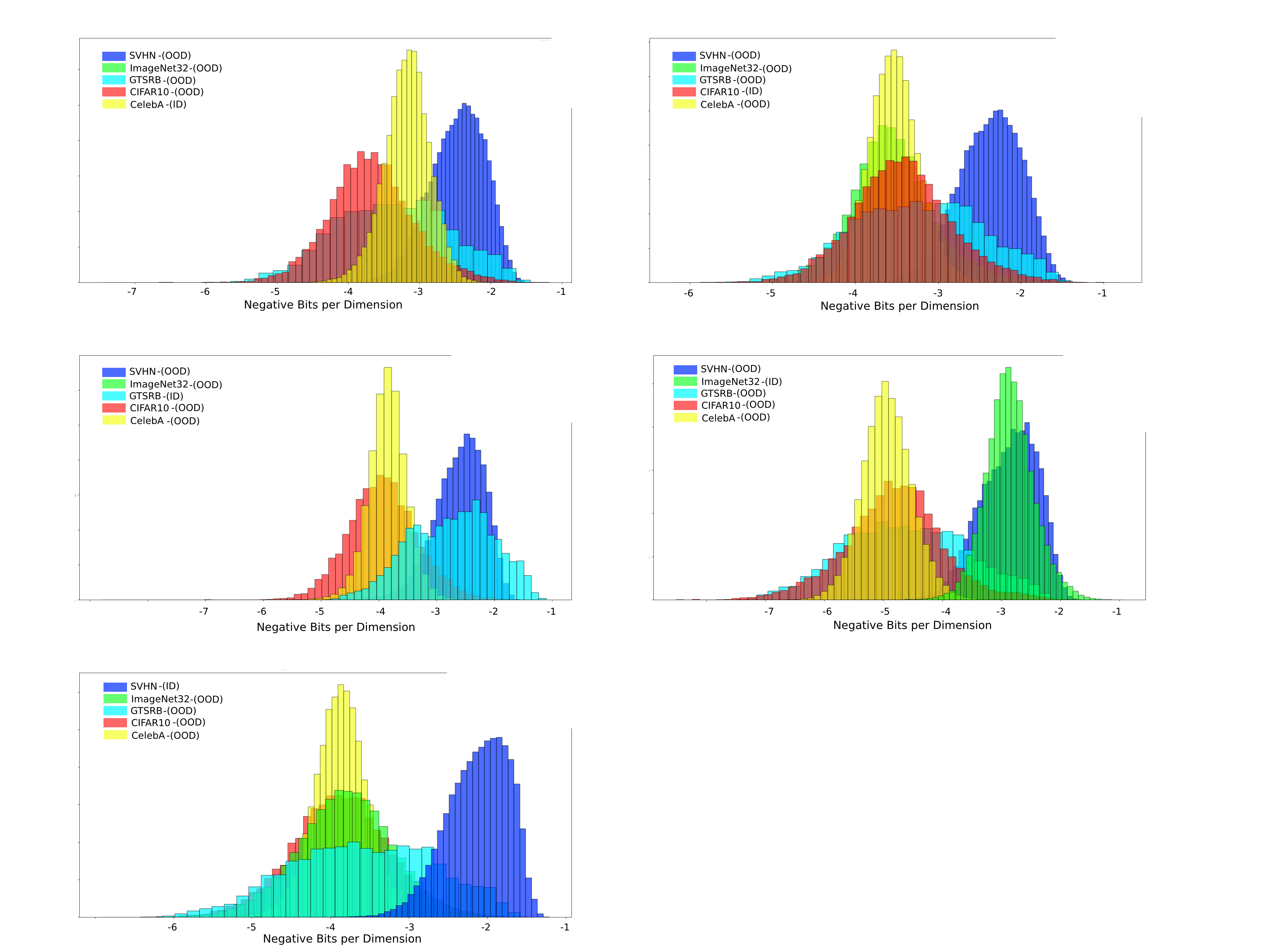}
    \caption{\textbf{\emph{Visualization of negative bits per dimension(BPD) failure as an OOD detection score for various ID-OOD dataset pairs.}}\\[0.5em]The figure illustrates the performance of the GLOW model when trained on a specific dataset (ID) and evaluated against four OOD datasets. In the first row (left), the model is trained on CelebA and tested on SVHN, ImageNet32, GTSRB, and CIFAR-10. Surprisingly, the model assigns higher likelihood (higher negative BPD) to OOD samples from SVHN than to its ID data, indicating a failure in OOD detection. Similarly, in the first row (right), the model trained on CIFAR-10 also demonstrates poor separation between ID and OOD datasets.
    In the second row (left), the model is trained on GTSRB, and in the second row (right), on ImageNet32. For all more complex datasets, the model assigns disproportionately lower NLL values to simpler OOD datasets like SVHN, signifying higher likelihood for OOD samples. This trend suggests that as the complexity of the ID dataset increases, the model struggles to differentiate OOD samples effectively.
    Finally, in the third row, the model is trained on SVHN. In this case, the GLOW model performs as expected, assigning frequently higher likelihood (higher negative BPD) to ID data. These observations highlight the inherent limitations of NLL as a stand-alone OOD criterion in reliably distinguishing between ID and OOD datasets, especially when dataset complexities vary significantly.}
    \label{fig:NLL_models}
\end{figure}

\section{Experimental Setups}
\label{sec:Experimental_Setups}

To train the three-channel colour datasets, the Glow model is applied using (CIFAR-10, GTSRB, ImageNet32, and SVHN) datasets with a batch size of 64 and a learning rate of 5e-4, for a total of 250 epochs. The network architecture included 3 blocks with 32 flow steps per block, 512 hidden channels, convolutional layers in affine coupling layer and inverse convolution for flow permutation.

For one-channel colour datasets, the model was trained on (FashionMNIST, MNIST, KMNIST and Omniglot) with a batch size of 128, using a learning rate of 1e-3 for 200 epochs. The architecture included one block with 10 flow steps, 1000 hidden channels, and linear layer in ACL layers, with a weight decay of 1e-4.

To calculate the OOD scores, the test split of each dataset was used to fit Gaussian to layer-wise logarithmic features. Additionally, 1000 random samples were selected to compute the AUROC results for each dataset.

\end{document}